\documentclass[sn-nature, twocolumn]{sn-jnl}

\usepackage{natbib} 

\usepackage{graphicx}%
\usepackage{multirow}%
\usepackage{amsmath,amssymb,amsfonts}%
\usepackage{amsthm}%
\usepackage{mathrsfs}%
\usepackage[title]{appendix}%
\usepackage{xcolor}%
\usepackage{textcomp}%
\usepackage{manyfoot}%
\usepackage{booktabs}%
\usepackage{algorithm}%
\usepackage{algorithmicx}%
\usepackage{algpseudocode}%
\usepackage{listings}%
\usepackage{geometry}
\usepackage{subcaption}
\usepackage{lipsum}
\usepackage[table]{xcolor}
\usepackage[dvipsnames]{xcolor}
\newgeometry{vmargin={15mm}, hmargin={17mm,17mm}} 

\theoremstyle{thmstyleone}%
\theoremstyle{thmstyletwo}%

\theoremstyle{thmstylethree}%

\begin{document}

\title{AutoLabs: Cognitive Multi-Agent Systems with Self-Correction for Autonomous Chemical Experimentation}

\author[]{\fnm{Gihan} \sur{Panapitiya*}}\email{gihan.panapitiya@pnnl.gov}

\author[]{\fnm{Emily} \sur{Saldanha}}\email{emily.saldanha@pnnl.gov}

\author[]{\fnm{Heather} \sur{Job}}\email{heather.job@pnnl.gov}

\author[]{\fnm{Olivia} \sur{Hess}}\email{olivia.hess@pnnl.gov}

\affil[]{\orgdiv{Pacific Northwest National Laboratory}, \orgname{Richland}, \orgaddress{\street{Washington}, \state{WA}, \country{USA}}}

\abstract{The automation of chemical research through self-driving laboratories (SDLs) promises to accelerate scientific discovery, yet the reliability and granular performance of the underlying AI agents remain critical, under-examined challenges. In this work, we introduce AutoLabs, a self-correcting, multi-agent architecture designed to autonomously translate natural-language instructions into executable protocols for a high-throughput liquid handler. The system engages users in dialogue, decomposes experimental goals into discrete tasks for specialized agents, performs tool-assisted stoichiometric calculations, and iteratively self-corrects its output before generating a hardware-ready file. We present a comprehensive evaluation framework featuring five benchmark experiments of increasing complexity, from simple sample preparation to multi-plate timed syntheses. Through a systematic ablation study of 20 agent configurations, we assess the impact of reasoning capacity, architectural design (single- vs. multi-agent), tool use, and self-correction mechanisms. Our results demonstrate that agent reasoning capacity is the most critical factor for success, reducing quantitative errors in chemical amounts (nRMSE) by over 85\% in complex tasks. When combined with a multi-agent architecture and iterative self-correction, AutoLabs achieves near-expert procedural accuracy (F1-score $>$ 0.89) on challenging multi-step syntheses. These findings establish a clear blueprint for developing robust and trustworthy AI partners for autonomous laboratories, highlighting the synergistic effects of modular design, advanced reasoning, and self-correction to ensure both performance and reliability in high-stakes scientific applications. Code: \url{https://github.com/pnnl/autolabs}}




\maketitle

\section{Introduction}\label{sec1}

Recent advances in artificial intelligence (AI)\cite{openai2024gpt4technicalreport, deepseekai2025, shinn_reflexion_2023, wang2024executablecodeactionselicit, schulman2017proximalpolicyoptimizationalgorithms,rafailov2024directpreferenceoptimizationlanguage,xu2025wizardlmempoweringlargepretrained, openai2025competitiveprogramminglargereasoning} and laboratory automation\cite{burger_mobile_2020,zhao_robotic_2023, manzano_autonomous_2022, slattery_automated_2024,jiang_autonomous_2023, greenaway_high-throughput_2018,lunt_modular_2024} are fundamentally transforming chemical research, promising to accelerate discovery through autonomous experimental design and execution.\cite{doi:10.1126/science.abc2986,doi:10.1126/science.aaa5414, manzano_autonomous_2022, greenaway_high-throughput_2018, granda_controlling_2018, he_algorithm-driven_2024, szymanski_autonomous_2023, zhu_automated_2024} Traditionally constrained by manual processes, low throughput, and complex hardware integration challenges, chemical experimentation is now poised for a revolutionary transformation driven by the emergence of agentic AI systems powered by large language models (LLMs)~\cite{gridach2025agenticaiscientificdiscovery,boiko_autonomous_2023, boiko2023emergentautonomousscientificresearch, campbell2025mdcrowautomatingmoleculardynamics, song_multiagent-driven_2025,vemprala_chatgpt_2024}. The ultimate vision of fully autonomous self-driving laboratories (SDLs)\cite{bennett_autonomous_2024, canty_science_2025, tom_self-driving_2024, macleod_self-driving_2020} encompasses the entire experimental lifecycle: from hypothesis generation and literature mining through experimental design, robotic execution, real-time monitoring, adaptive optimization, and results analysis and interpretation~\cite{yang2024largelanguagemodelsautomated, zhou2025hypothesispublicationcomprehensivesurvey, internagentteam2025internagentagentscientist, takagi2023autonomoushypothesisverificationlanguage, takagi2023autonomoushypothesisverificationlanguage, rabby2025iterativehypothesisgenerationscientific, newman2024arxivdigestablessynthesizingscientificliterature, chen2025chemminerlargelanguagemodel, song2024trialerrorexplorationbasedtrajectory}.  Recent systems have made significant strides toward this vision. The Coscientist system employs a GPT-4-based ``Planner'' to autonomously decompose synthesis goals, retrieve hardware documentation, and dispatch commands to cloud laboratories and liquid handlers~\cite{boiko_autonomous_2023}. Chemist-X uses retrieval-augmented LLM reasoning with on-the-fly program synthesis to optimize reaction conditions and control robotic wet-lab experiments~\cite{chen2025chemistxlargelanguagemodelempowered}. Similarly, 
ORGANA is an assistive robotic system that automates diverse chemistry experiments using LLMs for task interpretation, visual monitoring, and feedback. It supports parallel experiment scheduling and 3D object perception, and it has been demonstrated on tasks like solubility screening and electrochemical characterization, reducing chemists' workload while maintaining experiment quality\cite{DARVISH2025101897}.
Current automated chemical experiments span diverse applications including molecular synthesis, reaction optimization, catalyst discovery, materials screening, and functional characterization~\cite{dai_autonomous_2024, volk_alphaflow_2023, doi:10.1126/science.adi1407}.

Despite these advancements, the application of LLM-based agents in high-stakes scientific domains presents challenges that go beyond proof-of-concept demonstrations. 

The current evaluation methodologies for such agents tend to focus on holistic task success, often bypassing the granular examination of the agents’ underlying functionalities. This lack of systematic evaluation obscures an understanding of their failure modes, reliability, and areas for improvement. Addressing this gap is of paramount importance as these systems are deployed in domains where precision and trustworthiness are non-negotiable.

In this work, we present AutoLabs, an end-to-end LLM-based agentic system for automated chemical experiment design for Unchained Labs’ Big Kahuna high-throughput liquid handler\footnote{https://www.unchainedlabs.com/big-kahuna/}. 
While AutoLabs has been developed to generate hardware files for Big Kahuna, its experiment design capabilities extend beyond Big Kahuna, offering a versatile functionality that can be applied across diverse experimental setups. AutoLabs engages users in natural-language dialogue, plans experimental steps via a modular LLM-agent architecture, performs stoichiometric calculations through tool-calling, self-validates the generated protocol, and finally outputs the XML hardware file that drives the robot. We validate AutoLabs across five benchmark experiments of increasing complexity, ranging from calibration-sample preparation to multi-plate timed synthesis, and evaluate it under three levels of human-in-the-loop collaboration (none, non-expert, and expert). We design quantitative metrics to measure the correctness of the inclusion and ordering of the generated procedural steps as well as the accuracy of the final vial contents resulting from the procedure. Both these quantitative evaluations and a qualitative analysis by an expert systems engineer show that AutoLabs can achieve near-expert performance, substantially lowering the barrier for automated experimentation.

Our systematic evaluation reveals that the reasoning capacity of the LLM agents serves as the most critical factor for success, reducing numerical errors by over 85\% in complex tasks. When combined with a multi-agent architecture, and iterative self-correction, our system achieves near-expert procedural accuracy (F1-score $>$ 0.89) on challenging multi-step syntheses. These findings establish a clear blueprint for developing robust AI partners for autonomous laboratories, ensuring both capability and trust-worthiness as these systems are deployed in high-stakes scientific environments.

\section{Results}

\subsection{Agent Architecture}

Autolabs implements experiment design through a collaborative framework involving a multi-agent system and human users. The multi-agent system is implemented using LangGraph\footnote{https://www.langchain.com/langgraph}. LangGraph provides a graph-based architecture for complex AI workflows, where nodes represent processing steps (e.g., LLM invocations, tool integrations, or human inputs) and edges define data flow and control logic. This structure supports flexible workflows with branching, looping, and conditional execution, essential for advanced experimental design.

\begin{figure*}[!t]
    \centering
    \includegraphics[width=1\textwidth]{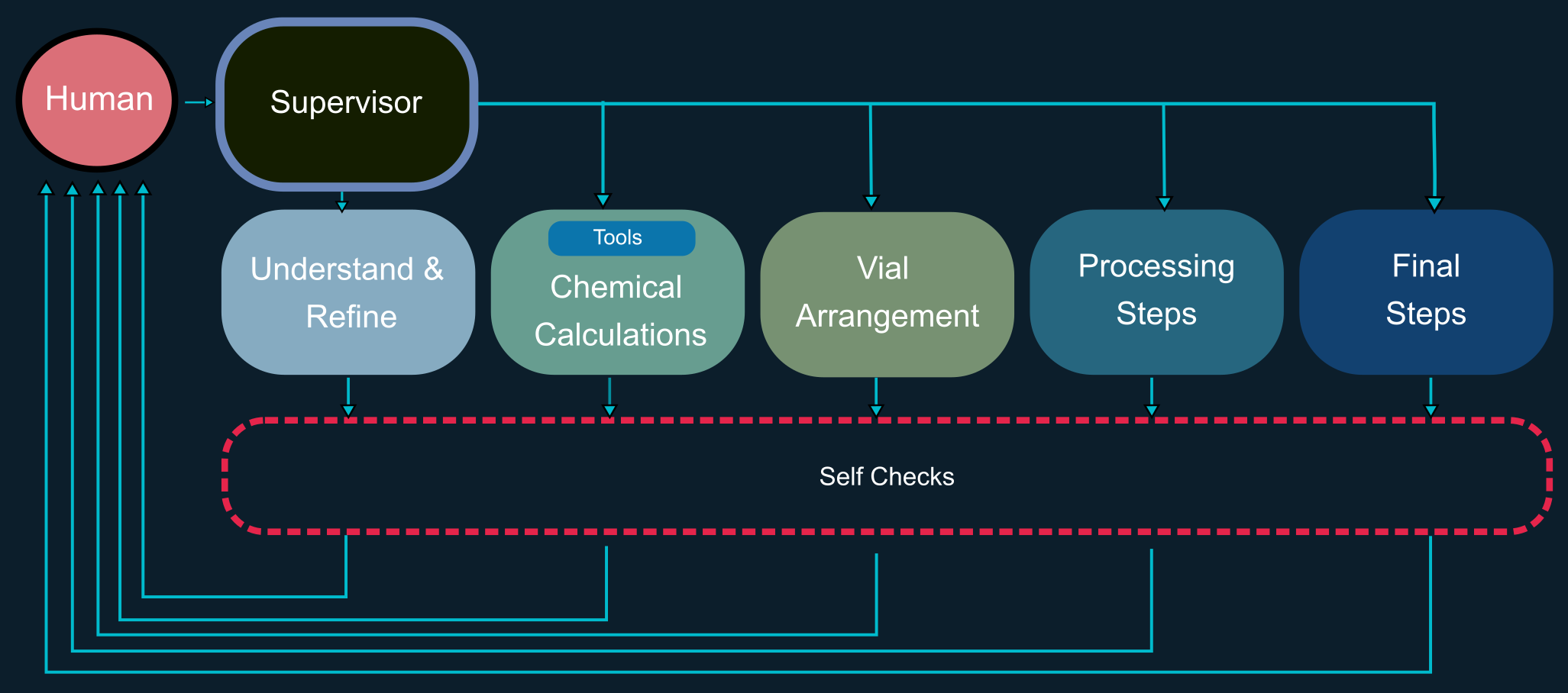}  
    \caption{Schematic of the multi-agent AutoLabs system.}
    \label{fig:agent}
\end{figure*}

Our multi-agent architecture consists of a supervisor agent orchestrating workflow among specialized sub-agents (Figure \ref{fig:agent}). An agent is an intelligent system powered by an LLM that can interpret natural language, have conversations, and autonomously perform tasks by combining reasoning, memory, planning, and tool use \cite{gridach2025agenticaiscientificdiscovery,
ghafarollahi2024sciagentsautomatingscientificdiscovery,sapkota2025aiagentsvsagentic, packer2024memgptllmsoperatingsystems, yuan2024easytoolenhancingllmbasedagents, yu2025chemtoolagentimpacttoolslanguage, yang2023gpt4toolsteachinglargelanguage, wu2025chematagentenhancingllmschemistry, zhu2025knowagentknowledgeaugmentedplanningllmbased, zhu2025healthflowselfevolvingaiagent, zhou2025greaterautonomymaterialsdiscovery, rivera2024conceptagentllmdrivenpreconditiongrounding, zhou2025scientistsexamprobingcognitive, yao2023reactsynergizingreasoningacting, xu2025advancingaiscientistunderstandingmultiagent, wei2025alignragleveragingcritiquelearning, wei2023chainofthoughtpromptingelicitsreasoning, lu2024aiscientistfullyautomated}. For the AutoLabs supervisor agent and sub-agents, we select from two GPT models~\cite{openai2024gpt4technicalreport}, GPT-4o and o3-mini. GPT-4o is a general-purpose model, well-suited for creative writing, conversational tasks, and multimedia processing whereas o3-mini specializes in reasoning-intensive tasks, leveraging chain-of-thought methodologies~\cite{NEURIPS2023_dfc310e8} for logical problem-solving and multi-step planning. 
Reasoning models\footnote{https://platform.openai.com/docs/guides/reasoning?api-mode=responses} like o3-mini are trained with reinforcement learning to handle complex problem-solving. 

The supervisor agent is initialized with a \textit{system prompt} that serves as a comprehensive guide to its functionality, outlining its capabilities, limitations, and requirements to ensure clarity in operation\cite{schulhoff2025promptreportsystematicsurvey, suzgun2024metapromptingenhancinglanguagemodels, Zhang_2025, jiang2023llmlinguacompressingpromptsaccelerated}. This system prompt provides essential information across several domains. It includes hardware specifications such as available vial sizes, array configurations, step types, and parameter limits, offering a detailed framework for the physical setup. Additionally, the prompt specifies syntax guidelines, outlining precise formatting rules for addition, processing, and quantitative steps, supplemented with examples for accurate implementation. It also incorporates chemical knowledge, embedding best practices such as capping volatile solvents and prioritizing the addition of solids over liquids to ensure safe and efficient procedures. Furthermore, reasoning protocols are integrated, providing structured methodologies for interpreting requirements, calculating quantities, and enabling effective user interaction. Together, these components enable the agent to operate within clearly defined parameters while delivering reliable and precise functionality. The full system prompt can be found in the section 1 of Supporting Information.

The experiment design process begins when a human user poses a question or provides a description of the experiment to be conducted. As illustrated in Figure \ref{fig:agent}, the human user interacts exclusively with the ``Supervisor'' agent, which serves as the central point of communication. Upon receiving the user’s input, the Supervisor evaluates the query and determines which of the five specialized sub-agents is best suited to address it. Each sub-agent is designed to handle a distinct aspect of experimental design, enabling a streamlined and efficient delegation of tasks.

\begin{figure}
    \centering
    \includegraphics[width=1\linewidth]{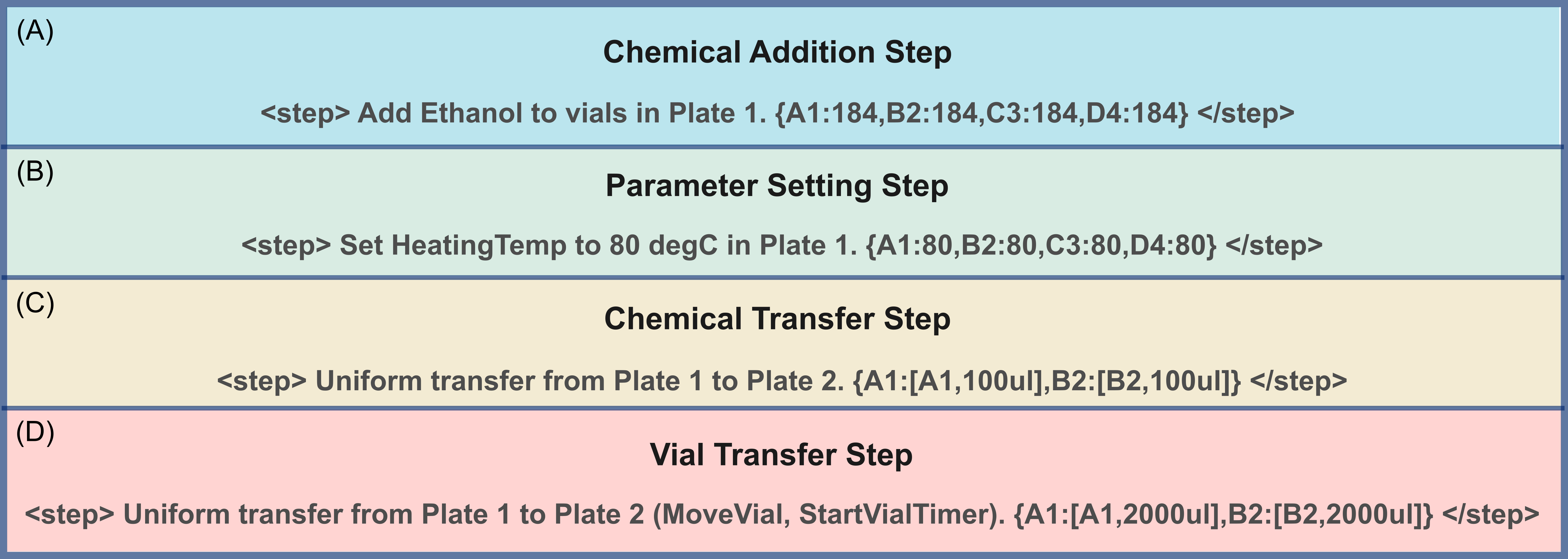}
    \caption{Format of an experiment step. (A) A regular chemical addition step. (B) A parameter setting step. (C) A chemical transfer step. (D) A vial transfer step: To differentiate it from a chemical transfer step, additional details are specified within brackets. For instance, the StartVialTimer instruction indicates initiating a timer to track the duration a vial remains in Plate 2. This ensures uniform timing for all vials at the transferred location prior to the next operation.}
    \label{fig:step_format}
\end{figure}

The sub-agents that are available to the Supervisor are listed here along with their roles:

\begin{itemize}
\item The \textbf{Understand and Refine} agent comprehends and refines experimental procedures, verifying understanding with the user and seeking clarification for ambiguities to ensure accurate task execution. 

\item The \textbf{Chemical Calculations} agent is equipped to perform precise stoichiometric calculations, leveraging advanced tool-calling capabilities. Four specialized tools implemented as Python functions have been defined to support its functionality: one calculates the volume of a chemical based on its name and weight; another determines the volume of a chemical given its name and the number of moles; a third computes the concentration of solutions; and the fourth calculates the required milligrams or microliters of two chemicals needed to prepare a solution with a specified molarity and volume. Additional details about these tools can be found in the Methods section. A description of the functions and their inputs and outputs are provided to the agent to assist in its decision-making about employing them.

\item The \textbf{Vial Arrangement} agent is tasked with organizing chemicals into vials to optimize the experimental workflow. It determines which chemicals should be allocated to specific vials, considering cases where multiple plates may be involved. This agent aims to minimize the number of steps required for adding chemicals.

\item The \textbf{Processing Steps} agent identifies necessary processing operations including heating, stirring, delays, and other procedural requirements for successful experimental execution. 

\item The \textbf{Final Steps} agent is responsible for organizing and sequencing procedural steps based on specified formatting requirements. These requirements include standardized tags and syntax for chemical names, processing parameters, and vial locations. The format of an AutoLabs experiment step is shown in Figure \ref{fig:step_format}. Although we have designated this special agent to focus on preparing the final steps, our tests revealed that, due to the nondeterministic nature of the LLM and the inclusion of final step instructions within the common system prompt, the agentic system can sometimes generate final steps at any stage. To address this behavior, the system has been designed to terminate the experiment generation process immediately upon the successful generation of the final steps, ensuring efficiency and preventing redundant outputs.
\end{itemize}

Once the final steps are produced as output using the specified formatting requirements, the generated steps are processed through the \textbf{Self-Checks} agent. This agent is specifically designed to validate the procedures against system requirements and user specifications through systematic verification of vial array sizing, output syntax, and procedure efficiency. Without this final verification step, we observed that long AI-human conversations could result in the LLM forgetting the instructions provided in the system prompt, especially given the challenge of complying with a large set of constraints and requirements.

We have implemented two types of step validation mechanisms in the Self-Checks agent: Guided and Unguided. In Guided Self-Checks, multiple specialized functions are employed to evaluate the steps according to each commonly observed type of LLM error. For example, there is one self-check function which asks the LLM to check whether the correct vial array has been used and another which asks the LLM to check that each chemical addition is specified in a separate step rather than adding multiple chemicals in one step. 

In Unguided Self Checks, the Self-Check agent is tasked to review the entire procedure for \textit{any} mistakes. For this mode, we provide the agent with the system prompt, the entire AI-Human conversation history, and the final steps along with instructions to modify the final steps if they do not adhere to the requirements. Rather than checking for each criteria and requirement one at a time, the final steps are evaluated holistically by the Self-Check agent. 
In Unguided mode, the Self-Checks agent is able to reprocess the steps multiple times up to a limit of 5 times until the agent believes that no errors are found in the steps.

After the self-checks are completed, users are invited to select appropriate hardware-related options, e.g., pipette tip size, for each step of the experiment. These options are tailored to the specific nature of the chemical being dispensed and the associated dispensing method. By leveraging a large language model (LLM) and domain knowledge provided by expert instrument engineers, recommendations for refining these options are suggested, ensuring a streamlined and accurate experiment design process. Brief descriptions of these tags are provided in the section 2 of Supporting Information. This tagging step concludes the input needed from the human-in-the-loop user, and AutoLabs next proceeds with translating the plain language experimental procedure that has been generated into instructions for the robotic system.

The final steps and options are input into a hardware file generation module to create the hardware file necessary for running experiments in Big Kahuna. This process is primarily carried out using rule-based coding. In existing literature, there are instances where LLMs have been employed for the task of hardware instruction generation. For example, in the CLAIRify tool~\cite{yoshikawa2023large}, a template is provided to the LLM, which then populates the template with values corresponding to the experiment. The filled template undergoes multiple revisions to ensure accuracy. Nonetheless, despite these refinement steps, LLMs may still fail to produce a fully accurate hardware file. 
Additionally, when dealing with extensive hardware files, such as those required for Big Kahuna, the generation process incurs a high token usage cost.
We observed that the structured nature of hardware files allows them to be reliably generated with perfect accuracy using rule-based coding. 
By specifying a structured but less-complex output format for the LLM procedure steps and using rule-based coding to translate that format to the hardware file syntax, we ensure precise file creation and mitigates errors. 
However, we do leverage LLMs for specific tasks within this pipeline, such as extracting chemical names from the experimental steps and determining chemical properties like molecular weight, density, and physical state.

\subsection{Evaluation Experiments}
To evaluate the AutoLabs framework we developed a set of test cases designed to probe typical use cases of the system of increasing complexity. For each test case, we manually create the correct experimental procedures to use for comparison with the procedures that are generated by the AutoLabs system.

These test cases are listed below.

\begin{enumerate}
    \item \textbf{Preparation of Calibration Samples}. \textit{Complexity}: low.\\
    This is a liquid sample preparation that would be used to calibrate an analytical instrument. This includes a small collection of samples (typically 5-10) used for anything that requires a set of known concentrations of analyte in solvent (i.e. HPLC, GC, UV, ICP). The overall design of this experiment has only 2-3 chemicals, a limited number of steps, and uses a single plate array. Even though it is a simple design, it tests the key potential stumbling blocks associated with running a stand-alone design, as well as ensures that the major functions of AutoLabs are performing as expected. 
        
    \textit{Experiment description}: Prepare a set of eight calibration samples for naphthalene in methanol. Total volume of each sample is 10 mL. Samples should have 5,10,15,20,25,30, 35 and 50 mg of naphthalene with the remainder being methanol. After preparation, the vials should be capped and then mixed via vortexing for 10 minutes. No heating required.\\

    \item \textbf{Preparation of Electrolyte Solutions}. \textit{Complexity}: low-medium.\\
    This is a step-up in complexity from Experiment 1 because it uses more chemicals and requires complex calculations. 

    \textit{Experiment description}: Prepare a set of electrolyte solutions, each consisting of a salt, solvent, and modifier.  I have three salts: lithium perchlorate, lithium tetrafluoroborate and lithium hexafluorophosphate.  I want to prepare 6 compositions with salt 1 and 2, loading the vials at 20 mg each, and 12 compositions with salt 3, loading half the vials with 20 mg and half with 50 mg.  The solvent is propylene carbonate and the modifier is ethylene carbonate, which will be dispensed as a solution of 1\% ethylene carbonate in propylene carbonate.  For each salt loading I want 500 uL to be the final volume, and the modifier concentration should be varied from 0\%, 0.2\%, 0.4\%, 0.6\%, 0.8\%, and 1.0\%.  After preparing these solutions, they should be heated to 40 deg for 30 minutes with stirring to ensure homogeneity.\\

    \item \textbf{Single Plate Synthesis}.  \textit{Complexity: Medium}. \\
    This is a synthesis experiment that involves a reactant, solvent, and an additive at different ratios, the requires heating to complete the reaction.

    \textit{Experiment description}: Perform a set of imine synthesis experiments using aqueous ammonia as the nitrogen source and solvent for the reaction.  Each reaction will be done in duplicate.  Each vial will contain two reactants: R1 and R2.  R1 is benzaldehyde at 0.5 mmol loading.  R2 will be one of 8 compounds, loaded at 0.75 mmol.  R2 chemicals are:  1-bromobutane, 1-iodobutane, 1-chlorobutane, 3-bromopropene, benzyl bromide, 3-bromobut-1-ene, 3-bromobut-2-ene, and 2-bromoethyl cyanide.  We want to see how the amount of ammonia affects the overall product yields, so the amount of water and 28\% aqueous ammonia solution should be calculated as to achieve ammonia loadings of 3M, 9M and 12M.  The total solution volume will be 1mL.  The samples will be headed at 60 degC overnight.\\ 
    
    \item \textbf{Multi-Plate Synthesis}. \textit{Complexity: High}.\\
    This is a synthesis experiment that includes post sample work up and requires the use of two arrays of vials. Running this test probes AutoLabs' ability to create multiple arrays which is a vital capability, since multi-array experiments are widely used.

    \textit{Experiment description}: Perform esterification reactions between an acid and an alcohol using the liquid catalyst sulfuric acid.  I would like to use the following alcohols: methanol, ethanol, propanol, and glycerol.  I have three acids:  acetic acid, propanoic acid, and benzoic acid.  Acetic acid should be combined with all four alcohols, but benzoic acid and propanoic acid should only be combined with methanol and ethanol.  I would like the molar ratio between the alcohol and acid to be examined at 0.5, 1.0 and 2.0, with the total Molarity of the acid and alcohol to be kept constant at 4 M.  The liquid catalyst should be set to 0.025 M, added as a 0.5M solution of sulfuric acid in water.  The remaining 2ml volume should be water.  After the solutions are prepared, they should be heated to 80 deg for 30 mins before cooling completely to 25C.  Once cooled, a portion of the sample can be diluted by a DF of 10 and transferred to HPLC vials (1mL total volume).  The HPLC samples should be vortexed for 20 minutes after being prepared. \\
    
    \item \textbf{Multi-Plate Synthesis With Timing}. \textit{Complexity: High}. \\
    This is a synthesis experiment that requires usage of timers and multiple plates. This is the most complicated type of experiment done on the Big Kahuna system.  

    In this experiment, we are keeping track of various reaction times as well as vial transfers between vial arrays.  

    \textit{Experiment description}: Perform time studies for the esterification reactions between acetic acid and 4 different alcohols (methanol, ethanol, propanol, and glycerol).  The molar ratio between acid and alcohol will be 1:1, with the total molarity of the acid and alcohol to be kept at a constant 4M. Sulfuric acid will act as the catalyst for this reaction.  We will target 0.025M in the solution, added as a 0.5M stock solution of sulfuric acid in water.  The remaining 2ml volume in each sample will be water.  Each mixture will be reacted at 80 degC at 6 time points (15, 30, 60, 90, 120, and 150 minutes).   
    
\end{enumerate}

\subsection{Evaluation Metrics}

We implement several quantitative metrics to probe the correctness of the procedure steps as well as the final chemical compositions resulting from the procedures. To evaluate the step correctness, we first need to match the AI generated steps with the ground truth procedure steps. There may be small differences in the chemical naming between the generated steps and the ground truth steps, so we use a fuzzy matching algorithm to identify the best match between the two sets of steps. We first converted each ground truth and generated step to the format ``Action $|$ Parameter $|$ Plate''. The $Action$ is one of Add, Set, Transfer, or Unknown. For an Add step, the $Parameter$ specifies the chemical name, while for a Set step the $Parameter$ specifies the processing parameter, e.g. HeatingTemp, Delay, StirRate, etc. The $Plate$ values lists which plate the action is performed for and is limited to be either ``Plate 1'' or ``Plate 2''. For each step in the generated procedures, we calculate a custom distance to each step in the ground truth procedures. This distance is set to infinity if the $Action$ or the $Plate$ do not match. For matching $Action$ and $Plate$ values, the distance is calculated using the Levenshtein distance~\cite{1966Levenshtein} of the $Parameter$ value. If the Levenshtein distance exceeds a maximum value of 5, the distance is set to a large value of $10^6$ to discourage matching of those steps. Finally, an optimal matching of steps between the two sets is performed using the linear sum assignment algorithm implemented in scipy~\footnote{\url{https://docs.scipy.org/doc/scipy/reference/generated/scipy.optimize.linear_sum_assignment.html}}. 

Once we have established the correspondence of steps between the generated and ground truth procedures accounting for small variations in chemical naming, we can compute metrics quantifying the success of the LLM at producing the correct steps. We calculate the \textit{precision} as the proportion of generated steps which are also present in the ground truth, the \textit{recall} as the proportion of ground truth steps which are present in the generated steps, and the \text{F1 score} as the harmonic mean of these two values. We calculate the \textit{Spearman correlation} by taking only the set of steps on which the generated and ground truth procedures agree and using the spearmanr implementation from scipy to compute the level of agreement between the step orderings.

The F1 metric and the Spearman metric don't take into account whether the correct chemical amounts are being added to the vials but only probe whether the correct sequence of steps are used. To evaluate the level of error in the vial contents, we again need to leverage our fuzzy matching approach to determine the best mapping between chemical names used in the ground truth procedures and the chemical names used in the generated procedures. Therefore, we apply the same Levenshtein distance and linear sum assignment procedure on the chemical names for all Add steps. For each chemical used in either version of the procedure, we compute the total amount of the chemical present in each vial by summing the quantities specified in each addition to each vial. If a chemical is used on the generated procedure but not the ground truth procedure or vice versa, we assign a quantity of zero for that chemical for the procedure where it is missing. 


We then compute the RMSE using each chemical-vial pair amount, $X_{c,v}$, as an observation. Because the total quantities of chemicals used differ significantly between the different experimental test cases, we compute a normalized RMSE (nRMSE) by dividing by range of the ground truth quantities, i.e. the maximum ground truth quantity value minus the minimum ground truth quantity value. 

\begin{equation}
nRMSE = \frac{\sqrt{\frac{1}{N_{chem-vial}}\sum_{c,v}{(X^{gen}_{c,v} - X^{GT}_{c,v})^2}}}{(\mathrm{max}(X^{GT}) - \mathrm{min}(X^{GT}))}
\label{eq:1}
\end{equation}

\begin{figure*}[!t]
    \centering
\includegraphics[width=1\textwidth]{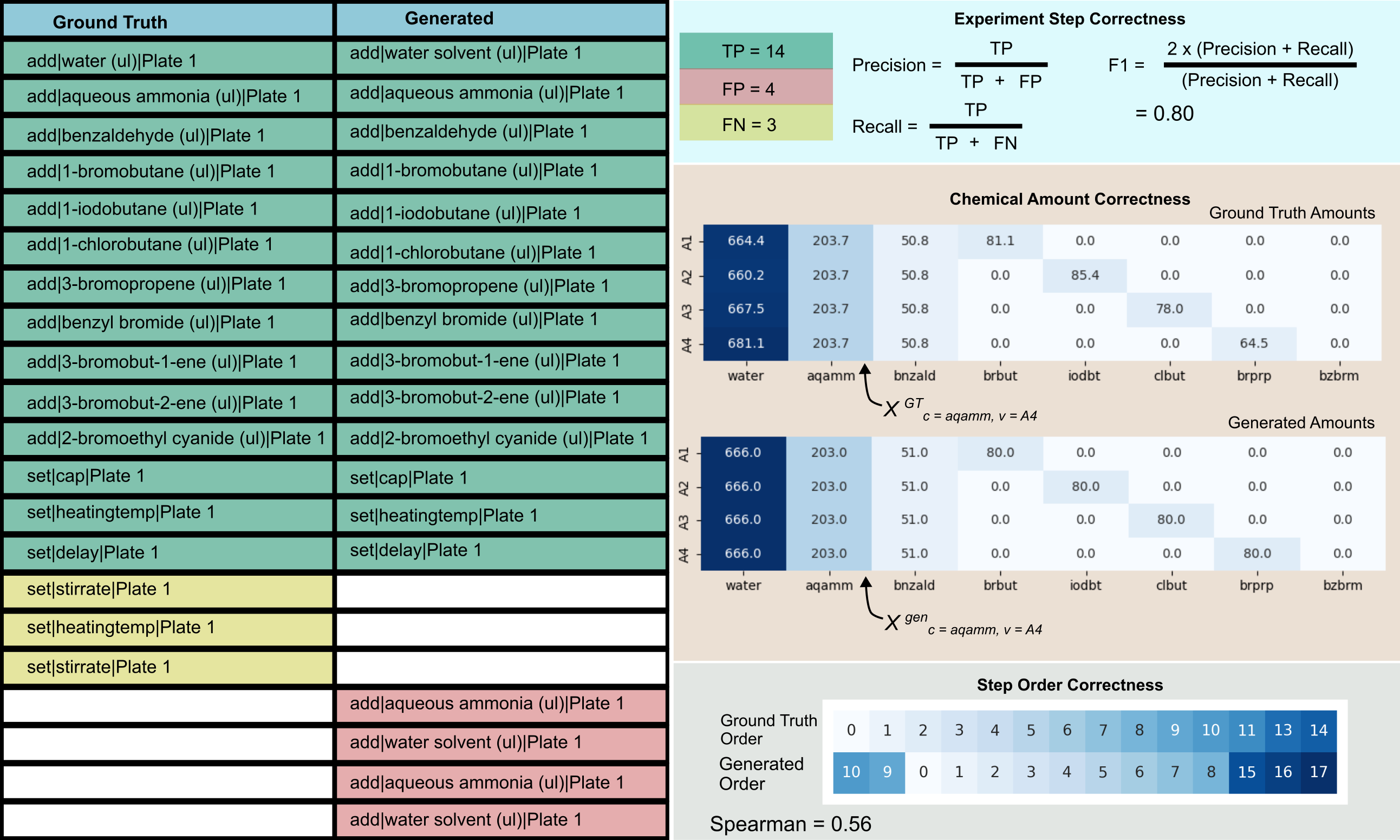}  

    \caption{Diagram illustrating the metrics computation for an example set of generated steps from Experiment 3. TP, FP and FN stand for True Positives, False Positives and False Negatives. $X_{c=aqamm, v=A4}^{GT}$ and $X_{c=aqamm, v=A4}^{gen}$ are specific examples of the values used in the computation of the nRMSE, defined in Equation \ref{eq:1}.}
    \label{fig:metrics}
\end{figure*}

We illustrate the computation of these metrics in Figure~\ref{fig:metrics} for an example generated procedure from Experiment 3. After finding the optimal mapping between ground truth and generated steps, we find that the generated steps have included extra additions for the aqueous ammonia and have omitted the stirring steps and the second heating temperature step to set the plate temperature back to room temperature. These matches and errors are used to compute the precision, recall, and F1 scores. Meanwhile, the matching steps in green are used to compare the ordering to compute the Spearman correlation. Because the generated steps add the water and ammonia at the end rather than the beginning like the ground truth, the overall Spearman correlation is reduced in this example. Finally, the full set of steps (along with the corresponding vial amounts provided in each step but not shown in the figure) is used to create matching arrays of vials contents which are are used to compute the nRMSE score.


\subsection{Evaluation Results and Discussion}

To assess the AutoLabs system, we first conducted a fully automated evaluation without human-in-the-loop collaboration to enable high throughput quantitative evaluation of the system's intrinsic knowledge and capabilities. In this fully automated setup, the AI-human conversation is initiated with a description of the experiment. Subsequent AI responses are automatically answered with the standardized input: ``Please use your best judgment and proceed''. This automated dialogue continues until the system detects final experiment steps in the AI response. If self-checks are enabled, the loop only terminates once the self-checks are completed. Throughout this process, the full chat history is preserved, ensuring contextual integrity. Once the termination condition is met, the experiment steps are extracted for calculating evaluation metrics.

\begin{figure*}[!htb]
    \centering
    \includegraphics[width=1\textwidth]{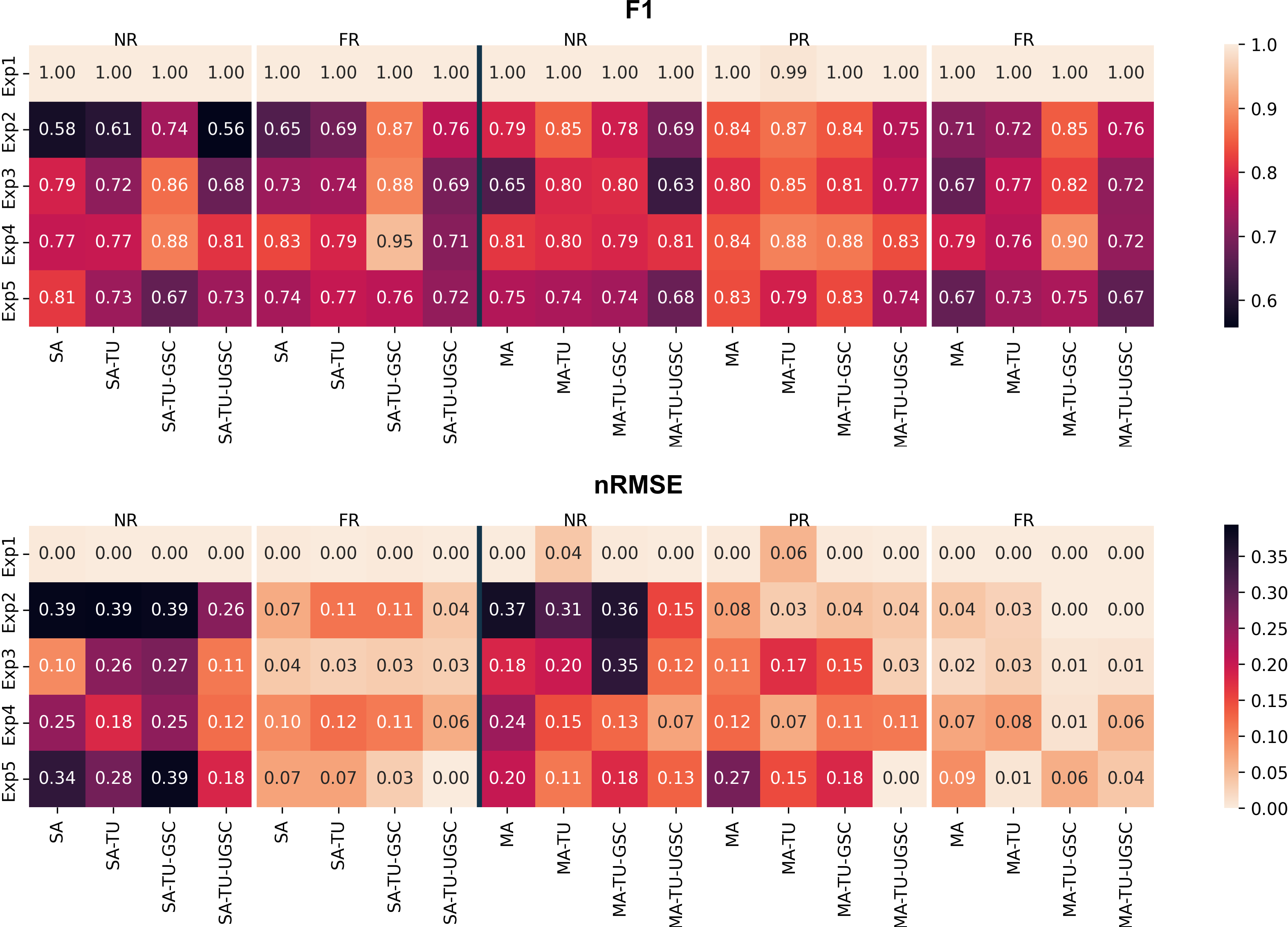}  
    \caption{Average F1 scores (top) and nRMSE scores (bottom) for each experiment across multiple architecture configurations in the fully automated evaluation mode.}
    \label{fig:all_results}
\end{figure*}

The complete set of evaluation results \textemdash ~gathered from experiments encompassing 20 distinct agentic architecture configurations \textemdash ~is illustrated in Figure \ref{fig:all_results}. Each configuration was executed 10 times, and the averaged metric outcomes are depicted for clarity and comparison.

The architectural configurations consisted of two primary frameworks: single-agent (SA) and multi-agent (MA). In the single-agent setup, only one agent interacts with the user, relying solely on a single system prompt to develop experimental procedures. Conversely, the multi-agent framework leverages a collaborative network of agents as described above. Details on the single-agent architecture is provided in the Methods section.

The configurations are also grouped into three categories based on the cognitive capacity of the LLM used by the agents:
No Reasoning (NR), Partial Reasoning (PR), and Full Reasoning (FR). NR configurations serve as a baseline, using the GPT-4o non-reasoning model for all agents. PR configurations take a hybrid approach, assigning the o3-mini reasoning model exclusively to the ``Understand and Refine'' agent while the others use GPT-4o. Finally, FR configurations equip all agents with the o3-mini reasoning model. Configurations were further compared based on their tool usage and self-checking mechanisms. For each cognitive capacity, a total of eight architectural variations are tested: 

\begin{itemize}
    \item \textbf{SA:} Single-agent without tool-calling or self-checking capabilities.
    \item \textbf{SA-TU:} Single-agent with tool-calling capability. 
    \item \textbf{SA-TU-GSC:} Single-agent with tool-calling and guided self-checking capability.
    \item \textbf{SA-TU-UGSC:} Single-agent with tool-calling and unguided self-checking capability.  
    \item \textbf{MA:} Mulit-agent without tool-calling or self-checking capabilities.
    \item \textbf{MA-TU:} Multi-agent with tool-calling capability.  
    \item \textbf{MA-TU-GSC:} Multi-agent with tool-calling and guided self-checking capability.
    \item \textbf{MA-TU-UGSC:} Multi-agent with tool-calling and unguided self-checking capability.

\end{itemize}

\begin{figure*}[!htb]
    \centering
    \includegraphics[width=1\textwidth]{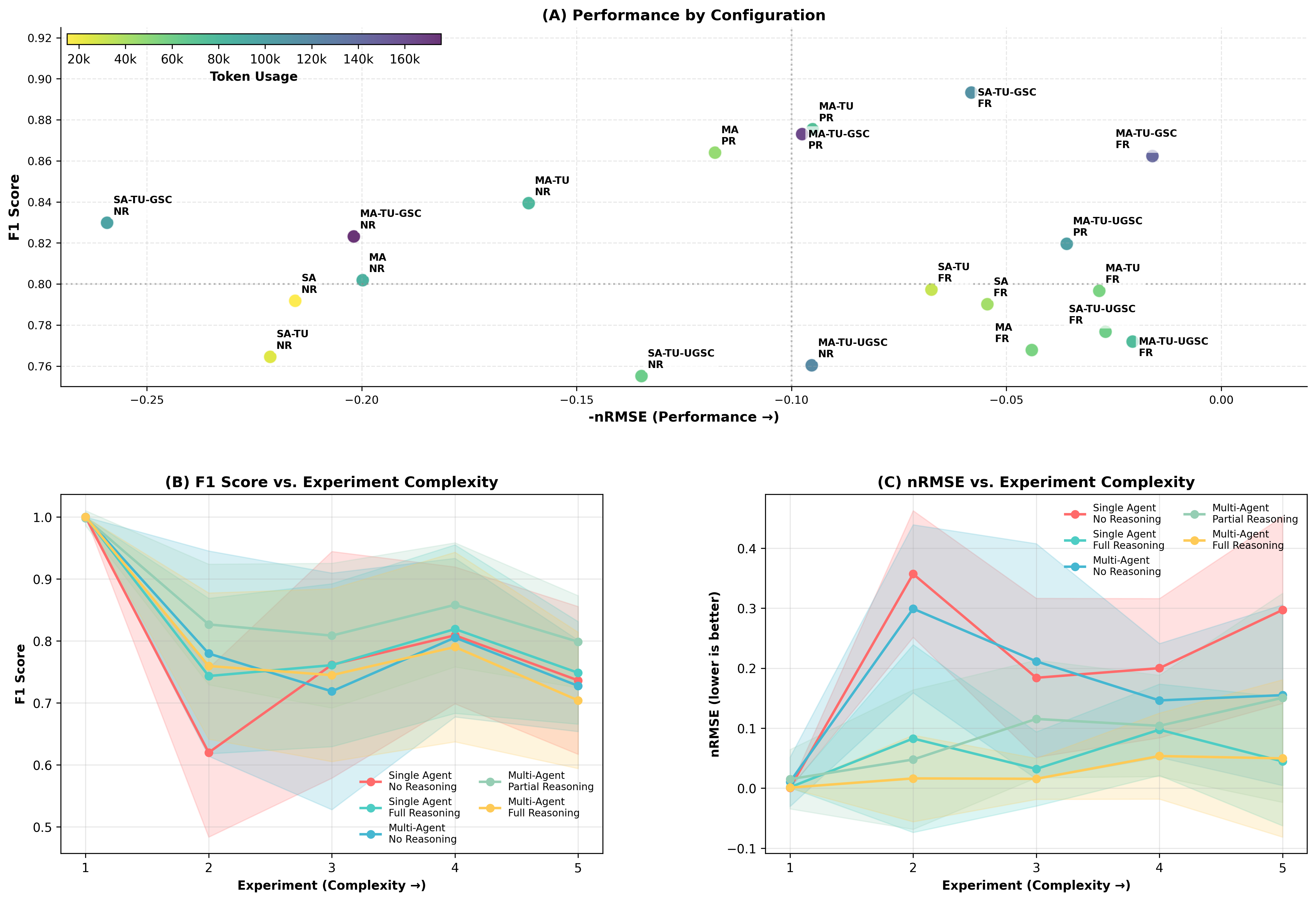}  
    \caption{Performance Trends Across Cognitive Configurations and Experiment Complexity.
(A) Average F1-score and nRMSE scores across all experiments for each architecture configuration. Color indicates average token usage the LLM.
(B) F1 Score vs. Experiment Complexity:
Line plots illustrate how step-generation accuracy (F1) changes with increasing experiment complexity (experiments 1–5). 
(C) nRMSE vs. Experiment Complexity: Line plots depict chemical amount error trends over experimental complexity.}
    \label{fig:config_F1_RMSE}
\end{figure*}

The averaged metrics across these 20 configurations are shown in Figure \ref{fig:config_F1_RMSE}(A). The results reveal that the full reasoning, multi-agent configuration equipped with tools and guided self-checks  achieves the lowest chemical amount error (nRMSE $\approx$ 0.03), while its' single-agent counterpart achieves the highest experiment step generation accuracy (F1  $\approx$ 0.87). To identify how experimental complexity impacts the accuracy, we show the average results for each experiment in Figures \ref{fig:config_F1_RMSE}(B) and \ref{fig:config_F1_RMSE}(C) for several architectural subsets. For procedure accuracy, multi-agent partial reasoning configurations maintain high F1-scores across all complexity levels. For chemical amount accuracy, full reasoning configurations consistently maintain low error rates as complexity increases, whereas configurations without reasoning exhibit higher error rates and variability.

In the following sections we discuss the effects of different architectural configurations on the accuracy of generating experiment designs.

\begin{figure*}[]
    \centering
    \includegraphics[width=1\textwidth]{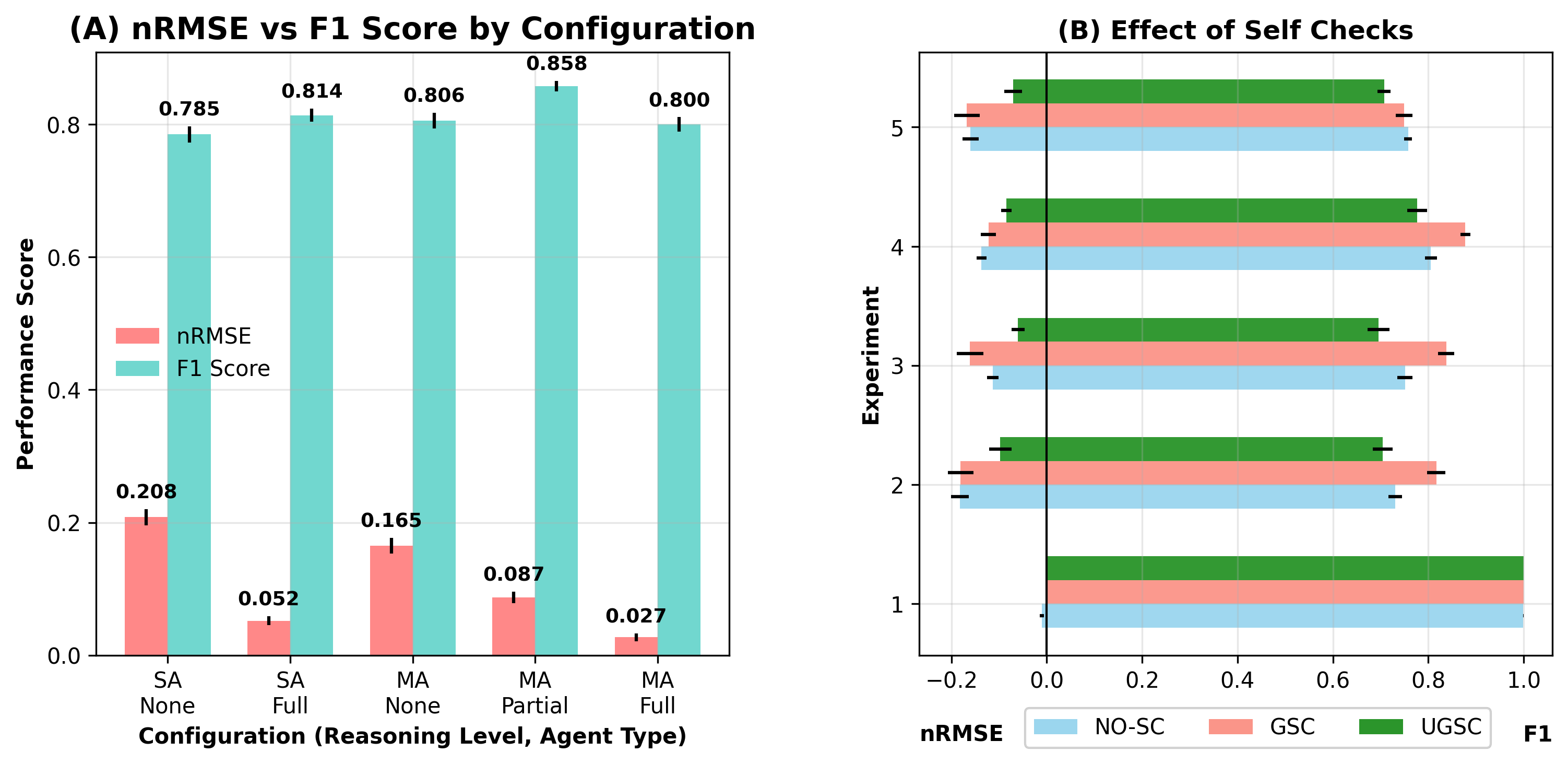}  
    
    \caption{Analysis of Cognitive Capacity and Architectural Effects on Performance. (A) Comparison of nRMSE and F1 scores across five system configurations. Error bars represent the standard error of the mean. Multi-agent systems with full reasoning achieve lowest nRMSE $\approx$ 0.027 while Multi-agent systems with partial reasoning achieve highest step generation accuracy (F1 $\approx$ 0.856). (B) Effect of Self-Checks. Left side shows negative nRMSE values, right side shows positive F1 scores. NO-SC: No Self-Checks, GSC: Guided Self-Checks, UGSC: Unguided Self-Checks.
    }
    \label{fig:trends}
\end{figure*}

\subsubsection{Reasoning Capacity is a Non-Negotiable Prerequisite for Quantitative Accuracy}

The most significant factor influencing performance across all experiments was the agent's reasoning capacity (Figure \ref{fig:all_results} and Figure \ref{fig:trends} A). While all architectures successfully generated the simple calibration protocol in Experiment 1, their performance diverged dramatically on tasks requiring conditional logic and complex calculations.

The impact of reasoning capacity was most starkly illustrated by nRMSE, which measures the accuracy of calculated chemical amounts. Across complex experiments (Experiments 2–5), models without reasoning capabilities (NR configurations) consistently exhibited high-error results, while reasoning-enabled models achieved substantial improvements in accuracy. When reasoning models were employed, single-agent systems demonstrated a 75\% decrease in nRMSE, while multi-agent systems exhibited an even greater enhancement of 83\% (with FR configurations). For instance, in Experiment 4, the non-reasoning SA configuration achieved an average nRMSE of 0.25, whereas the reasoning-enabled SA configuration demonstrated over 60\% better accuracy with an average nRMSE of 0.10. 

Experiment 2 exemplifies the critical importance of reasoning capabilities. The phrase ``ethylene carbonate, which will be dispensed as a solution of 1\% ethylene carbonate in propylene carbonate'' posed significant challenges for non-reasoning models to interpret correctly. Non reasoning models failed to understand that ethylene carbonate is added only via a 1\% volume-to-volume ethylene carbonate-in-propylene carbonate (EC-in-PC) stock. Instead, they erroneously treated the 1\% EC-in-PC stock as pure ethylene carbonate, leading to inaccurate ethylene carbonate calculations (refer to the section 3 of the Supporting Information for a detailed explanation of the calculations). In contrast, reasoning models successfully understood that 1\% stock contains 1 $\mu L$ ethylene carbonate per 100 $\mu L$ stock, resulting in impressive nRMSE reductions of 82\% and 89\% for single-agent and multi-agent systems, respectively.


The F1-score, which measures procedural correctness, also generally favored reasoning-enabled models, although the effect was less pronounced than for nRMSE. This pattern suggests that non-reasoning models can often identify correct procedural steps but fail critically in their quantitative parameterization. 


Beyond comparing fully reasoning (FR) and non-reasoning (NR) systems, we investigated partial reasoning (PR) strategies that employ reasoning models only for specific sub-agents within multi-agent architectures. This approach revealed important performance trade-offs that depend on the specific task requirements. For chemical amount generation, the FR strategy demonstrated clear superiority, achieving a substantially lower average nRMSE of 0.03 compared to 0.09 for the PR strategy. The PR strategy's limitations became particularly evident in Experiments 3-5, where errors stemmed from incorrect passing of arguments to tools, unit mismatches between calculated and ground truth values, and incomplete calculations. 

For protocol step generation, the PR strategy proved superior, achieving a higher average F1 score than the FR strategy (0.86 for PR vs. 0.80 for FR). Detailed analysis revealed that this advantage was driven primarily by superior performance in generating chemical addition steps (chemical addition step generation F1 scores: 0.89 for PR vs. 0.81 for FR). Both configurations demonstrated similar performance for parameter-setting steps, with average F1 scores of 0.80 for PR and 0.79 for FR. 

The higher F1 scores for chemical addition steps in the PR configuration can be attributed to its greater precision of 0.87, whereas the FR configuration suffered from lower precision of 0.74 due to a higher number of false positives. These false positives indicate that the FR configuration generated chemical addition steps that were absent from the ground truth. Manual inspection revealed these extraneous steps primarily involved the LLM distributing water additions across multiple steps—not chemical errors but rather protocol inefficiencies.

As illustrated in Table S6 in the Supporting Information, most PR configurations generated minimal or no superfluous water addition steps, with the notable exception of the PR configuration using guided self-checks. The FR configuration's tendency to generate these extra water addition steps indicates a concerning pattern where the reasoning model disregards explicit instructions provided in both the system prompt and self-check mechanisms, leading to unnecessarily complex protocols.



Examination of the factors contributing to the low F1 scores for parameter step generation in both PR and FR settings identified experiments 3 and 5 as the primary sources of reduced performance. An analysis of the precision and recall metrics related to parameter step generation (corresponding heatmaps are provided in the section 4 of Supporting Information) revealed that low recall values in experiments 3 and 5, combined with low precision values in experiment 5, were significant contributors to the diminished F1 scores.
These findings suggest a high prevalence of false negatives in experiments 3 and 5, where the systems consistently failed to generate critical parameter steps included in the ground truth. This issue appears to stem from incomplete robotic system specifications in the prompts, such as missing details regarding appropriate ranges for shaker speeds or incubation temperatures. Additionally, the low precision observed in experiment 5, indicative of an elevated number of false positives, suggests the generation of superfluous parameter steps that were absent in the ground truth.

To address these limitations, we explored using a Retrieval-Augmented Generation (RAG)\cite{gao2024retrievalaugmentedgenerationlargelanguage, fan2024surveyragmeetingllms} pipeline that enriched the ``Understand and Refine'' step with experimental procedures from three similar prior experiments. Testing this approach on MA (PR) and MA-TU-GSC (PR) configurations showed improvement in parameter step generation F1 scores for Experiment 5, though this improvement did not extend significantly to other experiments. For Experiment 5, the MA (PR) configuration exhibited an parameter generation F1 score increase from 0.49 to 0.63, while the MA-TU-GSC (PR) configuration showed an improvement from 0.57 to 0.65. However, significant enhancements in chemical generation step F1 scores were observed for Experiments 2, 3, and 4 when using RAG within the MA (PR) configurations. As a result, the overall F1 scores for Experiments 2 through 5 increased by 6\%, 12\%, 4\%, and 9\%, respectively. Refer to the section 6 of the Supporting Information for detailed F1 score tables.

These results demonstrate that while RAG-enhanced prompting shows promise as a strategy for improving automated protocol generation, its effectiveness varies depending on the specific task and experimental context. The targeted improvements in parameter step generation for Experiment 5 and the broader enhancements in chemical generation steps across multiple experiments highlight the potential of incorporating retrieval-based contextual information to address systematic deficiencies in LLM-generated protocols. However, the inconsistent improvements across all experimental conditions suggest that further refinement of the RAG approach, potentially including more comprehensive retrieval databases or adaptive retrieval strategies tailored to specific experimental requirements, may be necessary to achieve more uniform performance gains across diverse laboratory protocols.

\subsubsection{Guided Self-Checks Improve Procedural Accuracy. Unguided Self-Checks Improve Chemical Amount Accuracy.}

In comparing the results for different self-checking configurations, we found that guided and unguided self-checks play distinct roles in enhancing accuracy within experimental tasks. Guided self-checks significantly improve the accuracy of experiment step generation, while unguided self-checks are more effective in refining the accuracy of chemical amount calculations (Figure \ref{fig:trends} B).

Further analysis revealed notable benefits of guided self-checks, regardless of agent architecture (single-agent vs. multi-agent). 

\begin{figure*}[]
    \centering
    \includegraphics[width=1\textwidth]{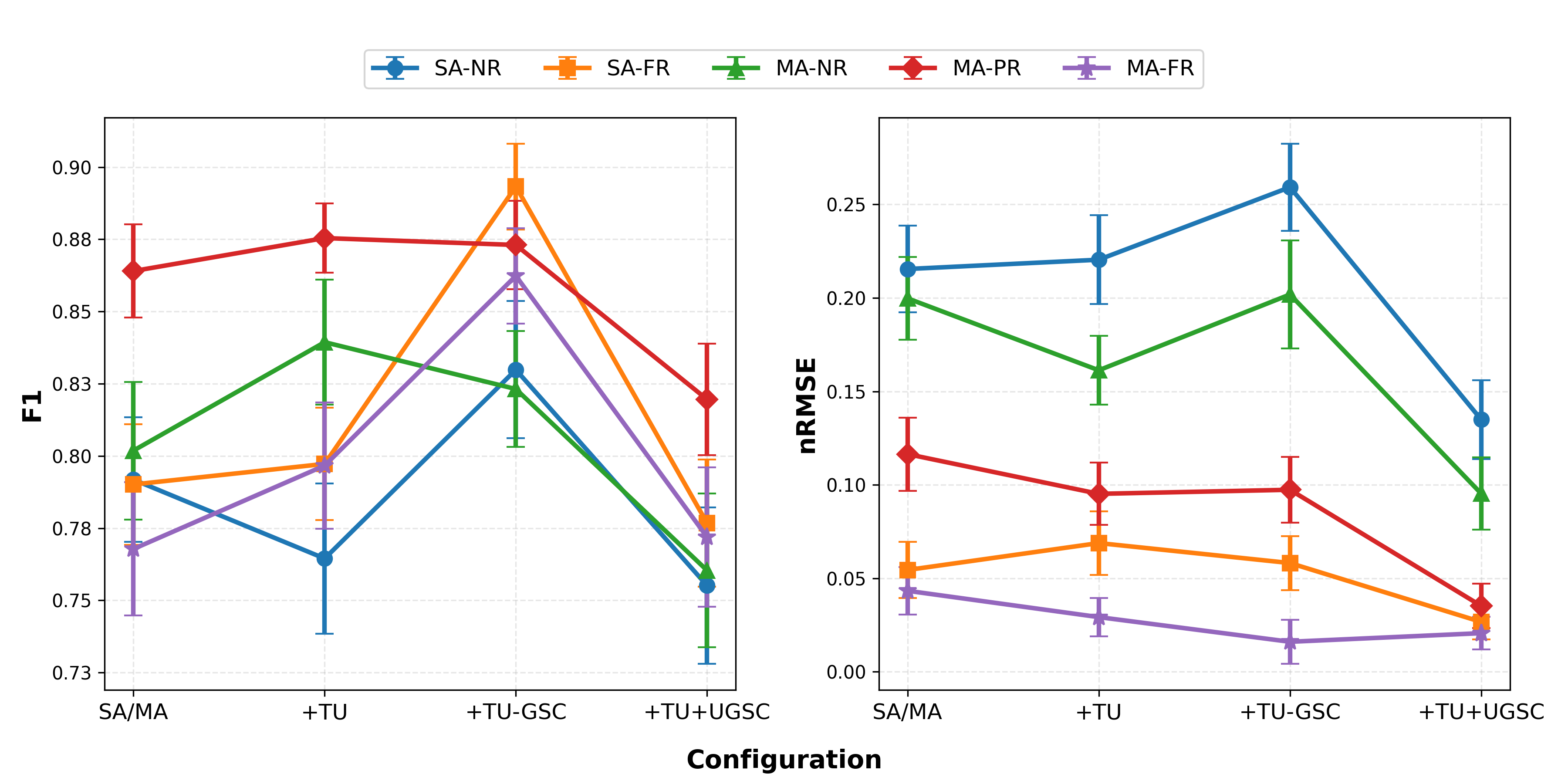}  
    \caption{Performance comparison across different agent configurations showing F1 scores (left) and normalized RMSE (nRMSE) values (right). 
The plot compares five agent types: Single-Agent No Reasoning (SA-NR), Single-Agent Full Reasoning (SA-FR), Multi-Agent No Reasoning (MA-NR), 
Multi-Agent Partial Reasoning (MA-PR), and Multi-Agent Full Reasoning (MA-FR) across four configurations: baseline (SA/MA), 
with tool usage (+TU), with tool usage and guided self-checks (+TU-GSC), and with tool usage and unguided self-checks (+TU+UGSC). 
Error bars represent standard error of the mean. The results demonstrate that self-checking mechanisms (particularly unguided self-checks) 
provide substantial performance improvements across all agent architectures, with multi-agent systems generally outperforming 
single-agent configurations in both metrics. Lines are provided as a visual guide only.}
    \label{fig:self-checks-trends}
\end{figure*}

As seen from Figure \ref{fig:config_F1_RMSE}, and highlighted in Figure \ref{fig:self-checks-trends} A, applying guided self-checks consistently improved the F1 scores of full-reasoning configurations for experimental step generation. This advantage extended even to single-agent configurations lacking reasoning models, highlighting the universal utility of guided self-checks. 

In contrast, the task of chemical amount generation showed a different trend where unguided self-checks were most effective (Figure \ref{fig:self-checks-trends} B). Compared to the baseline configuration, the unguided approach delivered universal improvements, reducing the nRMSE by 37\% for the non-reasoning single-agent model and by over 50\% for all others. Furthermore, for configurations without reasoning capacity (Single and Multi-agent NR), we observed a definitive drop in nRMSE when switching from guided to unguided checks, underscoring the superiority of the unguided method for these specific models. This pattern was also consistent across PR multi-agent configurations with the exception of Experiment 4. We should note however, that the unguided self checks utilize a reasoning model, thus the reasoning capacity of the model likely to has played a decisive role in the nRMSE reduction.

These findings underscore the nuanced role of self-checking mechanisms in improving procedural outcomes. Guided checks enhance procedural accuracy by offering a structured framework that mandates agents to validate their outputs against specific, predefined criteria. This approach targets key aspects such as procedural efficiency, unit consistency, correctness of calculations, and accurate use of critical parameters like timing and plate allocation, thereby significantly reducing common errors such as omissions and improper sequencing. On the other hand, the flexible, open-ended nature of unguided checks enables a more robust re-evaluation of quantitative reasoning, effectively ``lending" its reasoning capacity to improve numerical accuracy, even for agents that inherently lack it. This suggests that the optimal self-correction strategy is not one-size-fits-all, but rather is contingent on the nature of the task itself.

\subsubsection{Effect of Tool Usage} 

Our analysis revealed that full reasoning agents rarely opted to use the available tools. As a result, to assess the impact of tool integration, our study focused on agents with non-reasoning and partial-reasoning capabilities. To isolate the effects of the tools themselves, we compared agent performance with and without tool calls, while deliberately excluding self-correction mechanisms that could correct initial errors in the step generation.

\begin{figure*}[]
    \centering
  
    \includegraphics[width=1\textwidth]{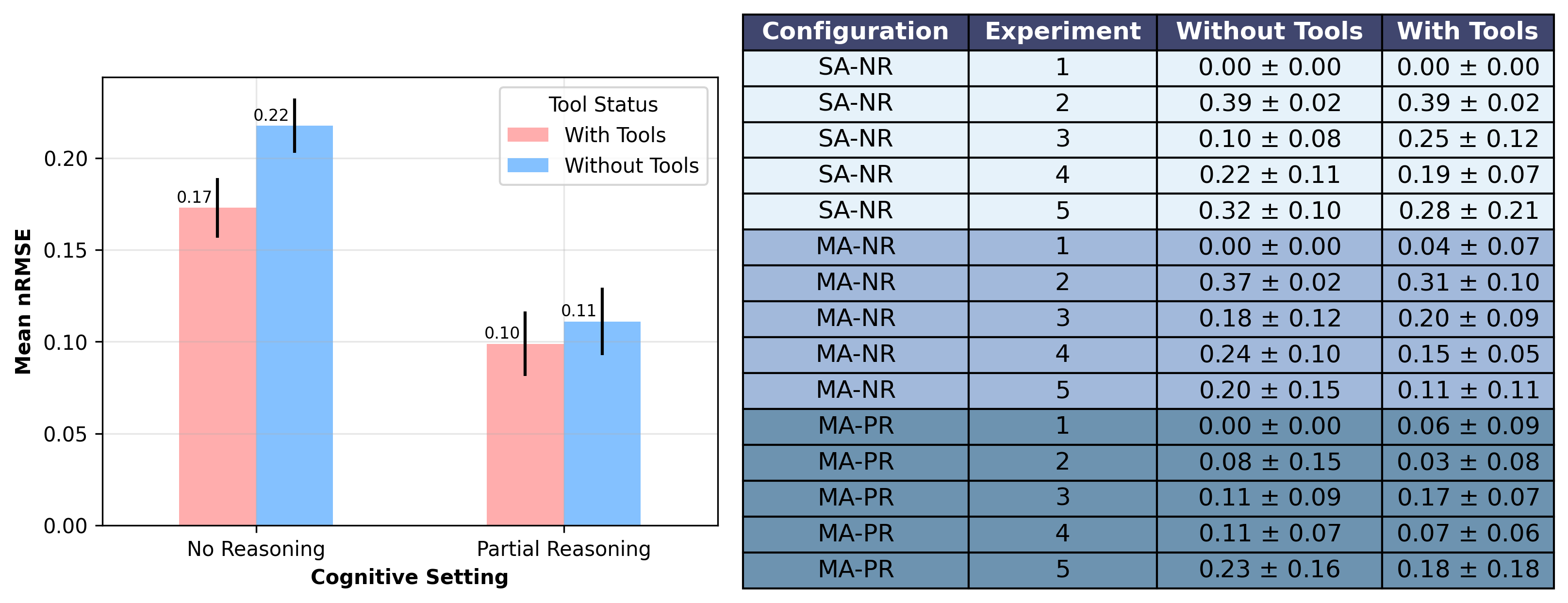}  
    \caption{Effect of tool usage on normalized RMSE (nRMSE) across different cognitive settings and agent configurations. Left: Bar plot shows the mean nRMSE for conditions with and without tool calls, separated by reasoning level (None, Partial). Right: Table summarizes the mean nRMSE for each configuration and experiment, comparing tool usage. Lower nRMSE values indicate better performance.}
    \label{fig:tools}
\end{figure*}

The integration of specialized calculation tools demonstrated that performance outcomes varied depending on the experimental context and system configuration (Figure \ref{fig:tools}, Right). In general, tool integration improved the accuracy of chemical quantity calculations in esterification experiments (Experiments 4 and 5) when self-check mechanisms were absent, compared to scenarios where the tools were not utilized. These findings align with expectations, as the find\_chemical\_amounts\_in\_a\_solution tool was specifically designed to support calculations relevant to such experiments. Consequently, the LLM exhibited less ambiguity in selecting this specialized tool over more generalized alternatives.

However, in Experiment 3, the use of tools had a detrimental effect. Further analysis identified incorrect calculations of the amount of 28\% ammonia as a key source of errors in the design of Experiment 3. Our toolset includes a function (find\_the\_concentration\_of\_n\_percent\_solution) specifically designed to calculate the molarity of percentage-based chemical solutions by extracting the weight percentage from the chemical name via an LLM query. For the 28\% ammonia solution, we observed that the agent incorrectly passed "ammonia" as the chemical name parameter instead of the full specification, "28\% ammonia." This error led to the use of an incorrect molarity for the 28\% ammonia solution, impacting the accuracy of the experiment.

These results suggest that while the integration of specialized calculation tools can significantly enhance performance in some contexts, their overall utility depends on the LLM's ability to use them accurately. This highlights the need for further research into optimizing tool integration, particularly by examining the impact of the number of available tools and the quality of their descriptions on the LLM's ability to navigate and apply them effectively.

\subsubsection{Agent's Path Analysis}

\begin{figure*}[]
    \centering
    \includegraphics[width=1\textwidth]{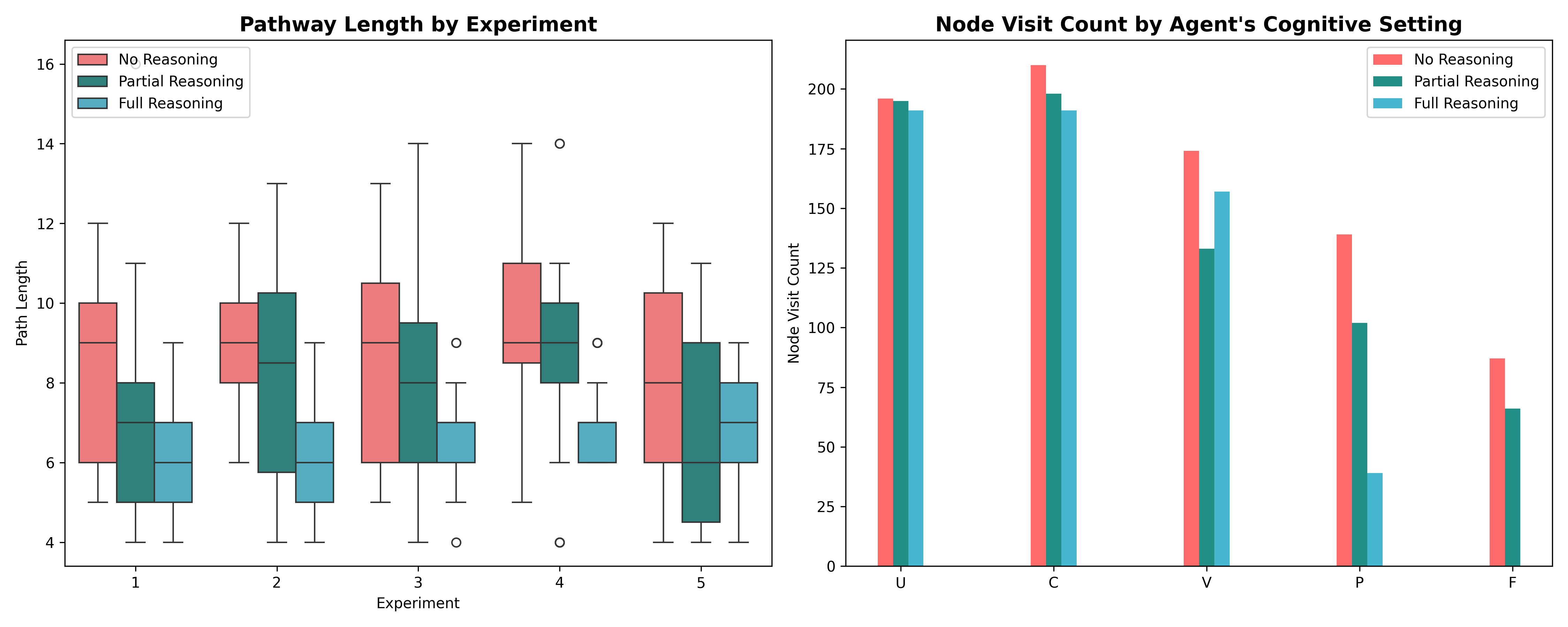}  
    \caption{(a) Path length variation by experiment and agent's cognition level. (b)  The total count of node visits by agents at different cognitive levels, aggregated over all 10 repetitions of the experiments. U: Understand and Refine, C: Chemical Calculations, V: Vial Arrangement, P: Processing Steps, F: Final Steps.}
    \label{fig:path}
\end{figure*}

We also conducted an analysis of pathway length (Figure \ref{fig:path}), which is defined as the number of agent nodes in the pathway from the start to the completion of the task. Path lengths were observed to vary depending on the level of reasoning and the specific experiment. Agents employing full reasoning typically utilize shorter paths, while agents without reasoning tend to follow longer paths.
Another notable observation is the reduced reliance on processing and final steps nodes as the cognitive level increases. In fact, agents with full reasoning exhibited only one instance of utilizing the final steps node. This highlights the critical role of both calculation and vial arrangement nodes as pivotal elements within the experimental design workflow.

\subsubsection{The Accuracy-Cost Relationship}

\begin{figure*}[]
    \centering
    \includegraphics[width=1\textwidth]{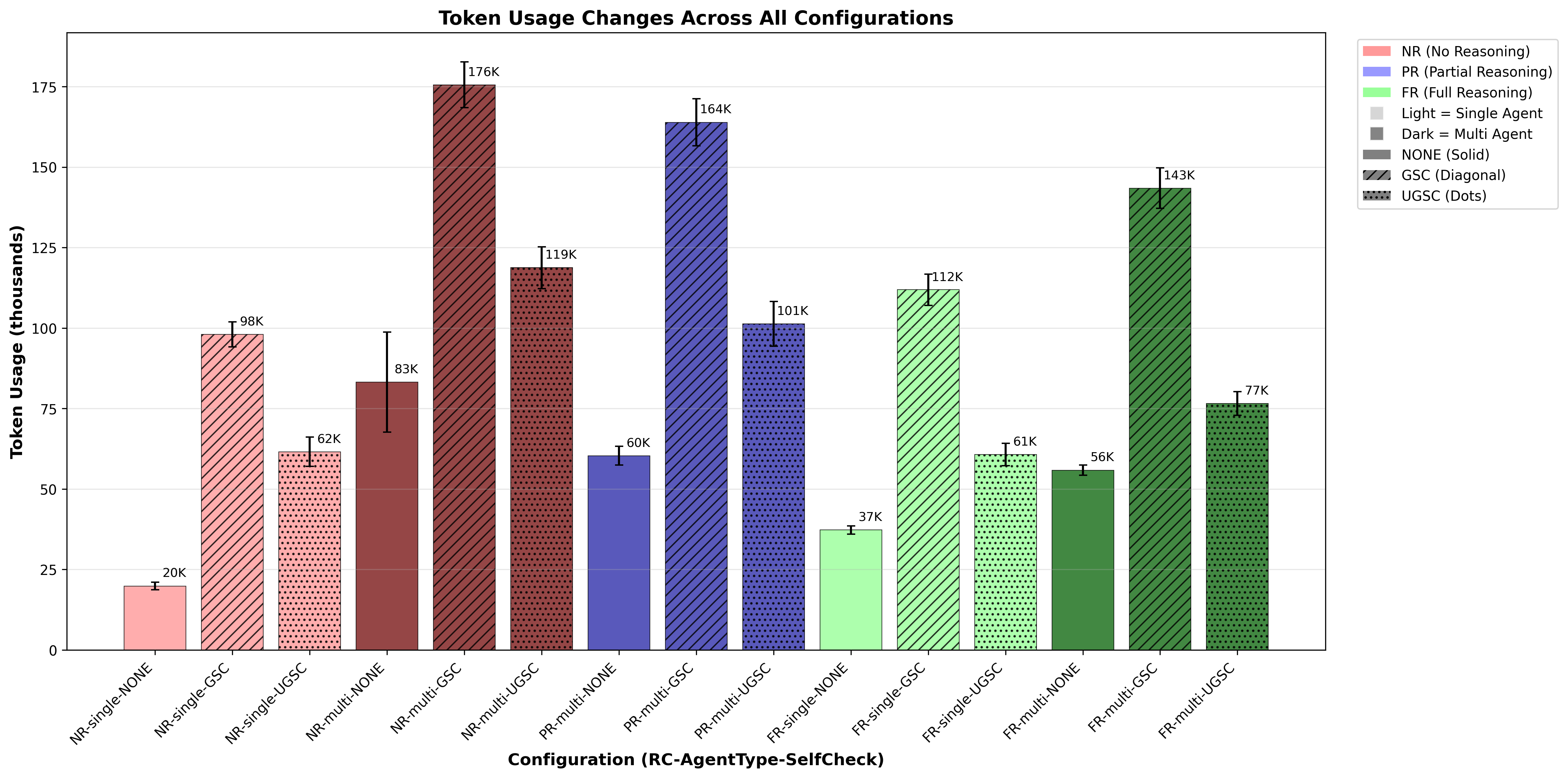}  
    \caption{Average total token usage of different configurations for each experiment.}
    
    \label{fig:efficiency}
\end{figure*}

The highest-performing agents incurred substantial computational overhead (Figure \ref{fig:efficiency}). Specifically, the integration of self-correction mechanisms significantly increased token consumption.

In the single-agent architecture, we observed a clear correlation between computational investment and performance outcomes: greater cognitive effort yields globally better results. Implementing full reasoning capabilities over the baseline no-reasoning approach increased token costs by 24\%. This computational investment delivered improvements on both key metrics: protocol accuracy showed modest gains, with average F1 scores across all experiments improving from 0.785 to 0.814, while chemical amount prediction error (nRMSE) decreased dramatically by 75\%.

Analysis of token usage and accuracy across multi-agent configurations reveals that enhanced reasoning capacity improves design correctness without increasing token consumption. For instance, the PR and FR configurations achieved 48\% and 83\% improvements respectively in chemical amount generation accuracy despite using significantly fewer tokens than the NR configuration. We attribute the FR configuration's token efficiency to its ability to determine the necessary steps more directly, a conclusion supported by its significantly shorter agent path lengths compared to the NR model.

\subsubsection{Correlation Analysis of Generated Steps with Ground Truth}

\begin{figure*}[htbp]
\centering
\includegraphics[width=1\linewidth]{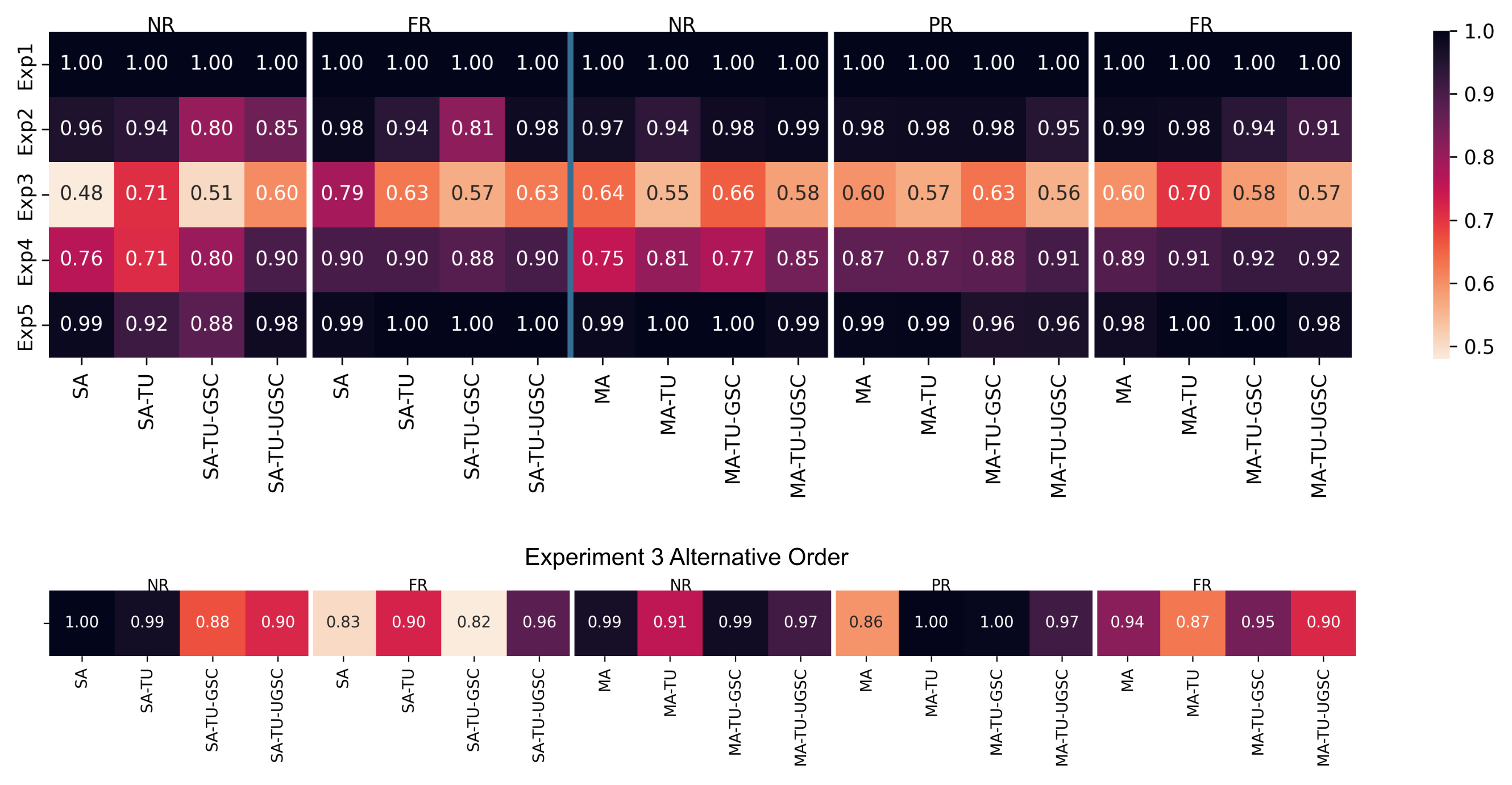}
\caption{Spearman correlation scores measuring the agreement between the generated and ground-truth experimental step order. The top panel shows results for the original step ordering, while the bottom panel shows results for a valid alternative ordering for experiment 3. This alternative ordering is provided in the section 7 of the Supporting Information.}
\label{fig:spearman}
\end{figure*}

Thus far, our evaluation of the various AutoLabs configurations has primarily emphasized the F1-score and nRMSE metrics as key indicators of performance. In contrast, the Spearman correlation has been treated as a secondary metric, largely because strict adherence to step ordering is often unnecessary for ensuring procedural correctness. While achieving high Spearman correlation scores is not our primary focus, the procedural steps generated by AutoLabs consistently exhibit impressive Spearman correlation values relative to the ground truth across nearly all experiments—with the notable exception of Experiment 3 (Figure \ref{fig:spearman} top panel).

Upon closer examination, the anomaly in Experiment 3 was traced to discrepancies in the sequence of chemical additions between the ground truth and the generated steps. Specifically, the ground truth specifies adding water and ammonia initially, while the system-generated steps prioritize the addition of other chemicals before introducing water and ammonia. This deviation, however, is scientifically valid within the context of the experiment. To further investigate, we recalculated the Spearman correlation values after revising the addition order in the ground truth procedure—adjusting it to reflect the system-generated sequence, where other chemicals are added first, followed by water and ammonia. The corresponding Spearman correlation values are presented at the bottom of Figure \ref{fig:spearman}. This adjustment resulted in a substantial improvement in the average Spearman correlation, elevating the average score from 0.61 to 0.93 across all the configurations. 

This observation highlights several critical aspects regarding the evaluation and validation of procedural steps generated by AutoLabs. First, AutoLabs consistently demonstrates robust capabilities in procedure generation, delivering sequences that align closely with expected outcomes, as evidenced by high Spearman correlation values across most experiments. Even in cases of divergence, such as Experiment 3, the procedurally generated steps remain scientifically valid, underscoring the system's flexibility in adapting to domain-specific variations where multiple acceptable solutions may exist. These findings, however, underscore the limitations of rigid reliance on ground truth as a sole benchmark, as it represents only one of many valid procedural orders.

To address these limitations, it is vital to adopt a more context-aware validation framework that accounts for domain-specific nuances. Consulting domain experts in cases of deviation is essential to assess the scientific credibility of such divergences, thereby bridging gaps between automated systems and traditional norms. This collaborative approach ensures that procedural outputs generated by AutoLabs not only adhere to technical accuracy but also align with practical expectations and domain-specific correctness.

\subsubsection*{The Optimal Agent} 
The top-performing agent across all the experiments when both procedural step generation and chemical amount generation accuracy were considered  was MA-TU-GSC with Full Reasoning (FR). 
This configuration successfully combines problem decomposition (multi-agent), and guided self correction, leading to a high F1 scores and lowest numerical errors (Figure \ref{fig:config_F1_RMSE}).

\subsection{Impact of Human-In-the-Loop}

Following our fully automated evaluation of AutoLabs, which was used to identify the architectural configuration with the strongest intrinsic capabilities, we next performed evaluation of AutoLabs in several human-in-the-loop collaboration modes, one with non-expert interactions and one with expert interactions.

The goal of conducting non-expert evaluations was to assess the accuracy of experimental designs generated by AutoLabs when used by individuals with minimal knowledge of chemical experiments. These evaluations were conducted by three data scientists/software engineers each of whom generated the procedures for each experiment. In this mode, the users were instructed to answer the LLM's questions about the experiment but not to proactively correct mistakes made by the LLM. This is designed to probe the impact of correcting any ambiguities in the initial experimental description while still relying on the LLM for most of the domain knowledge. 

\begin{figure*}[htb]
    \centering
    \includegraphics[width=1\textwidth]{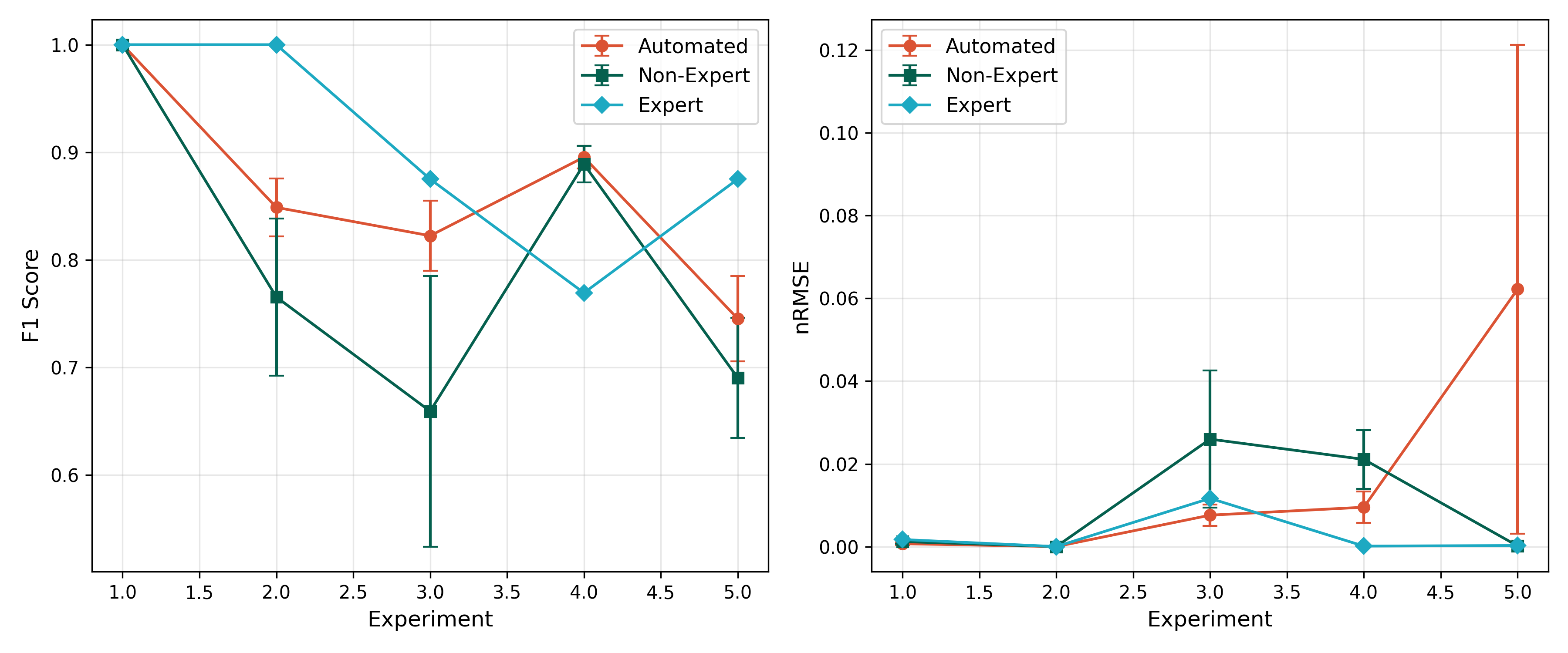}  
    \caption{Comparison of F1 score and normalized nRMSE across five chemical experiments, evaluated by three methods: automated metrics, non-expert human evaluators, and expert human evaluator. Error bars indicate standard error of the mean for automated and non-expert results.}
    \label{fig:human_evals}
\end{figure*}

In this context, we did not anticipate significant deviations from the fully automated evaluations. However, our findings shown in Figure \ref{fig:human_evals} indicate that for experiments categorized as simple to moderately difficult, protocol generation was more precise when handled by the fully automated model compared to the input provided by non-experts. This highlights the advantages of allowing the large language model (LLM) to rely on its own judgment rather than depending on potentially unclear or ambiguous responses to its questions from non-expert users. 

In our expert evaluations, the evaluations were performed by a systems engineer with deep expertise in robotic automation and implementing experimental procedures in the Big Kahuna system. The expert was instructed to work with the AI agent to achieve the most accurate experimental outcome possible by answering the agent's clarifying questions and providing corrective feedback for any procedural or calculational errors. In this mode, we find that expert collaboration leads to significant improvements in F1-score and nRMSE across most experiments relative to non-expert and fully automated modes.
However, we find that even a user with deep domain knowledge failed to catch certain errors generated by AutoLabs. These overlooked mistakes were often related to procedural processing steps—such as setting stir rates or capping plates—which can have inherent ambiguity and are easily missed when an expert is focused on the core chemical additions. For instance, in several experiments, the expert did not notice when AutoLabs omitted critical stir rate commands (Experiment 3), failed to assign cap values and stir rates to the correct plates (Experiment 4), or incorrectly assigned a `StirRate 0' command and added a superfluous delay step (Experiment 5).

Ultimately, this study demonstrates that while human-in-the-loop collaboration can be highly effective, it is not a universal solution for all errors. The significant performance gains from expert interaction confirm the value of a human partner in resolving high-level conceptual and calculational mistakes. Conversely, the degradation in performance with non-expert input and the procedural oversights by the expert both reinforce the distinct advantages of a well-prompted, fully automated system. This suggests a powerful synergy for future systems: combining expert oversight for strategic experimental design with rigorous, automated validation for procedural details could create a more robust and reliable workflow than either approach could achieve alone.

\subsection{Qualitative Evaluation}

To complement the quantitative metrics of our high-throughput, automated procedures, a qualitative evaluation was conducted by the expert systems engineer who generated the expert evaluation described in the previous section. The objective was for the expert to assess the performance and usability of the AutoLab system across five distinct experiments through a collaborative interaction model. As noted above, the expert was instructed to work with the AI agent to achieve the most accurate experimental outcome possible by answering the agent's clarifying questions and providing corrective feedback for any procedural or calculational errors. The evaluation revealed a spectrum of performance, from flawless execution to instances requiring significant user intervention, highlighting both the system's strengths and areas for future development.

According to the expert's assessment, the system's performance in Experiment 1 was flawless, with the entire workflow executing smoothly without requiring any user intervention. The expert found that Experiment 2 was also managed successfully, noting the system's proficiency in accurately calculating the final vial compositions. A minor limitation was identified when the system was unable to consolidate the addition of lithium hexafluorophosphate into a single step, despite repeated user prompts. The expert observed this issue to be reagent-specific, as the system correctly consolidated other compounds upon request.

The expert noted that this proficiency with intricate designs was further demonstrated in Experiment 3, where the system correctly mapped an even more complex layout onto the specified plate in its first attempt. While the system was observed to initially omit density from its calculations, the expert highlighted that it correctly flagged this omission in the output table—a critical alert that, in the expert's opinion, a non-expert might overlook. Following the expert's intervention, the system successfully recalculated all values with unique, density-adjusted volumes. However, procedural inconsistencies were also recorded, specifically in the sporadic application of post-heating cooling and stirring operations.

In Experiment 4, the expert was again impressed by the system's ability to process a complex design. It was observed that the system proactively sought clarification on the physical state of starting materials and preferred calculation methods. Although it initially erred by not adding a solid component first, the expert found that it corrected the sequence flawlessly after a single request, and the calculations were praised for their accuracy.

The final evaluation, Experiment 5, revealed what the expert described as a notable divergence between the system's planning and execution. The initial conceptualization of Vial Timers was deemed excellent, but the implementation was only partially successful. The expert found that the system erroneously inserted an unintended delay that rendered the timers ineffective. While this step was removed upon request, other shortcomings were seen to persist, including the frequent omission of mixing steps and the failure to cool the reaction plate post-run. This final omission can be attributed to the lack of an explicit cooling instruction in the prompt.

Based on these observations, the expert concluded that the system's reliability is highly dependent on the user's ability to craft comprehensive and detailed prompts. This led to the suggestion that integrating instrument-specific standard operating procedures (SOPs), potentially via methods like RAG, is essential for designing more robust and truly autonomous automation workflows with generative AI.

\section{Conclusion}

In this work, we introduced AutoLabs, a self-correcting multi-agent architecture designed to automate chemical experiment design, and conducted a comprehensive, systematic evaluation of its performance. Our investigation, spanning 20 distinct agent configurations across five benchmark experiments, has yielded critical insights into the design of reliable AI agents for autonomous laboratories. We have demonstrated that the reasoning capacity of the underlying LLM is the single most important determinant of success, particularly for ensuring the quantitative accuracy of chemical preparations, where it reduced numerical errors by over 85\%. While reasoning is paramount, our findings also confirm the significant benefits of a multi-agent architecture for decomposing complex problems and a dedicated self-correction loop for refining outputs. We identified a nuanced but crucial division of labor for self-correction: guided checks are most effective at improving procedural correctness (F1-score), whereas unguided, holistic reviews by a reasoning model excel at correcting numerical inaccuracies (nRMSE).

Furthermore, our study of human-AI collaboration revealed a powerful synergy. While expert oversight is invaluable for correcting high-level conceptual and calculational errors, even experts can overlook minute procedural details like setting stir rates or capping plates. A fully automated, well-prompted system with robust self-correction excels at these detail-oriented verifications. This suggests an optimal workflow where human experts guide the overall strategy while the AI agent reliably manages the granular implementation and validation, creating a partnership more robust than either could achieve alone.

The qualitative analysis and performance limitations observed also highlight key areas for future work. The system's occasional procedural omissions and sensitivity to prompt detail underscore the need to embed deeper domain knowledge and enhance its adaptive capabilities. One promising avenue is integrating instrument-specific Standard Operating Procedures (SOPs) via Retrieval-Augmented Generation (RAG) to make the system more robust. Beyond static retrieval, future work should explore automated prompt optimization, allowing the system to dynamically refine its internal prompts to elicit more accurate and reliable responses from the LLM. Integration of memory is another critical step; enabling agents to retain context from previous steps and learn from past experiments would prevent repeated errors and build a cumulative, experience-based understanding of laboratory protocols. A more ambitious direction involves implementing an agent evolution mechanism, where the multi-agent architecture itself can adapt over time. By testing different agent configurations and collaboration strategies on new tasks, the system could learn to dynamically assemble the most effective team for a given scientific problem, truly mimicking an expert research group's growth and adaptation. Further research is also needed to improve the reliability of tool usage and to generalize the AutoLabs framework beyond the Big Kahuna platform to other laboratory hardware. Ultimately, the principles and methodologies established by AutoLabs pave the way for a new generation of AI-driven scientific tools—not merely as automated assistants, but as reliable, self-correcting partners in the complex and creative process of scientific discovery.

\section*{Declarations}

\section{Methods}

\subsection{Single-Agent Architecture}

Unlike the multi-agent system that distributes tasks across specialized modules, the single-agent system consolidates the entire experimental design workflow within a single supervisory function. The key distinction between the system prompts for single-agent and multi-agent systems lies in the inclusion of descriptions for various agents within the multi-agent system prompt. This architecture has access to the same chemistry calculation tools used by the multi-agent system  and to the guided and unguided self-check mechanisms.

The single-agent workflow begins with system prompt injection containing domain-specific instructions for experimental protocol generation. When tools are enabled, the agent iteratively invokes chemistry-specific functions until all calculations are resolved. Upon detection of finalized experimental steps (indicated by <final-steps> tags), the system optionally applies self-checks procedures before terminating the workflow.

\subsection{Tools}

Here we provide descriptions for the Python function-based tools that were provided the ``Chemical Calculations'' agents.

\subsubsection*{get\_chem\_volume} Calculates the volume of a chemical by dividing chemical's weight by it's density. The inputs to this function are the name of the of chemical and the weight. The molecular density is found using an LLM using the chemical name.

\subsubsection*{find\_the\_volume\_corresponding\_to\_moles}
This function is designed to calculate the volume of a chemical, given its name and the number of moles in SI units. It operates by first retrieving the molecular weight  of the chemical using an LLM call. The function then multiplies this molecular weight by the number of moles to calculate the total weight of the chemical. Next, it retrieves the chemical's molecular density also using an LLM. Once the density is determined, the function calculates the chemical's volume (V) by dividing its total weight by the density.

\subsubsection*{find\_the\_concentration\_of\_n\_percent\_solution} 
This tool is designed to calculate the molar concentration of a chemical in an n\% (weight/volume) solution. It takes the chemical name as input and uses external methods to retrieve the weight percentage, molecular weight, and molecular density of the specified chemical. The first step involves querying a model to determine the weight percentage of the chemical. Next, the molecular weight and molecular density are obtained by invoking a LLM. Using these values, the function computes the total mass of the chemical present in 1 liter of solution, assuming a density-based calculation of mass. The function then calculates the mass of the chemical in an n\% solution by multiplying the total mass by the weight percentage and dividing by 100. Based on this mass and the molecular weight, the number of moles of the chemical is determined. Finally, the function calculates the molar concentration (moles per liter) of the solution.

\subsubsection*{find\_chemical\_amounts\_in\_a\_solution}

This function calculates the quantities of two chemicals required to prepare a solution with a specified total molarity, molar ratio, and volume. This function is useful when dealing with solutions where the molar ratio between the two chemicals can be clearly determined. It accepts the total molarity of the solution, the molar ratio between the two chemicals, the names of both chemicals, and the volume of the solution as inputs. 

First, the function calculates the molarity of each chemical based on the given molar ratio. These values represent the molar concentrations of the two chemicals. Next, the total number of moles of each chemical is calculated by multiplying the molarity of the respective chemical by the volume of the solution. Using the molecular weight of each chemical, obtained via a call to a LLM, the mass of each chemical in grams is calculated by multiplying the number of moles by the molecular weight. The volumes are calculated by dividing the mass by the substance's density, and these volumes are returned as the required amount.

\subsection{Self Checks}

\subsubsection*{Guided}
In the Guided Self-Checks specialized refinement functions are sequentially invoked, each of which checks a specific aspect of the experiment protocol, such as step efficiency, unit consistency, timing, plate usage, solvent specification, transfer details, and chemical addition structure. Here is a list of these verification steps:

\begin{itemize}
\item \textbf{refine\_efficiency}: Ensures that each input chemical is added using as few steps as possible, streamlining the experimental protocol for efficiency and minimizing unnecessary actions.

\item \textbf{refine\_units}: Verifies that the units for chemical additions are consistent and appropriate: solid additions should be in milligrams (mg), and liquid additions in microliters ($\mu l$), supporting accurate and standardized reporting.

\item \textbf{refine\_delays}:Checks that sufficient delays are included for steps involving heating, stirring, or vortexing, ensuring that physical processes are given adequate time for completion and reproducibility.

\item \textbf{refine\_plates}:Validates the use of correct plate formats and array dimensions based on vial sizes, confirming that chemicals are assigned to the proper locations and that vial indexing is consistent throughout the protocol.

\item \textbf{refine\_solvents}: Reviews the specification and usage of solvents in the protocol, ensuring that solvent volumes are calculated correctly and that solvent choices are appropriate for the chemicals and reactions involved.

\item \textbf{refine\_transfer}: Examines the details of chemical transfer steps between plates, confirming that transfers are described clearly and accurately, with proper tracking of source and destination containers.

\item \textbf{refine\_additions}: Assesses the structure and tagging of chemical addition steps, ensuring that each addition is annotated with the correct tags (such as ``Powder'' for solids or ``SyringePump''/``PDT'' for liquids) and that optional tags are applied as needed for clarity and automation.
\end{itemize}

\subsubsection*{Unguided}
Contrary to the Guided Self-Checks, for the Unguided Self-Checks a single validation prompt is used to guide the review of generated experiment steps. The prompt includes user specifications, system capabilities, and common error checks, which is then passed to the validation agent (o3-mini with medium reasoning) for evaluation.

\subsection{Hardware File Generation}

We first conducted a detailed manual analysis of the structure of Big Kahuna hardware files, which were generated during prior experiments. These files are organized into three primary sections: (1) the chemicals used in the experiment, (2) parameters related to instrument settings, and (3) the experimental steps necessary to achieve the desired objectives.
The information required to generate these hardware files is derived from experiment steps produced by AutoLabs and the user-selected tags for each step. We leverage LLM calls to extract essential details, including chemical and parameter names, the number of plates utilized in the experiment, the physical state of the chemicals (solid or liquid), and other key properties such as molecular weight and density. This extracted information is then used to systematically and accurately complete the data across all three sections of the hardware file.

\section*{Declarations}

\subsubsection*{Funding}
This research was funded by the Generative AI for Science, Energy, and Security Science \& Technology Investment under the Laboratory Directed Research and Development Program at Pacific Northwest National Laboratory (PNNL), a multiprogram national laboratory operated by Battelle for the U.S. Department of Energy. This work was also supported by the Center for AI and Center for Continuum Computing at PNNL. 

\subsubsection*{Conflict of interest}
There are no conflicts of interest to declare
\subsubsection*{Consent for publication}
Not applicable
\subsubsection*{Data availability }

\subsubsection*{Code availability }
The code is available at \url{https://github.com/pnnl/autolabs}
\subsubsection*{Author contribution}
G.P. conceived the idea, led the project, contributed to developing agent architecture, analyzed the results and contributed to writing the manuscript. E.S. provided project administration, contributed to implementation of agent architecture and evaluation methods, and contributing to writing and editing the manuscript. H.J. designed evaluation experiments, provided expert knowledge on the operation of Big Kahuna and performed expert evaluations. O.H. contributed in implementing agent capabilities. G.P, E.S. and O.H. performed non-expert evaluations.

\subsubsection*{Competing interests}
The Authors declare no competing interests.

\bibliography{sn-bibliography}

\end{document}


\title{Supporting Information - AutoLabs: Cognitive Multi-Agent Systems with Self-Correction for Autonomous Chemical Experimentation}


\author*[1]{\fnm{Gihan} \sur{Panapitiya}}\email{gihan.panapitiya@pnnl.gov}

\author[1]{\fnm{Emily} \sur{Saldanha}}\email{emily.saldanha@pnnl.gov}

\author[1]{\fnm{Heather} \sur{Job}}\email{heather.job@pnnl.gov}

\author[1]{\fnm{Olivia} \sur{Hess}}\email{olivia.hess@pnnl.gov}

\affil*[1]{\orgdiv{Pacific Northwest National Laboratory}, \city{Richland},  \state{WA}, \postcode{99354}, \country{United States}}

\maketitle

\section{System prompt}
This section provides the detailed system prompt used to guide the multi-agent system in generating experimental steps for the AutoLabs robotic platform. It outlines the capabilities of the robot, the expected format for generating steps, and the roles of different agents within the system.

\begin{lstlisting}
AutoLabs is a robotic system for automated chemical experiment execution. 

You are the supervisor agent of a multi-agent system tasked with generating the steps to complete a given experiment. 
Your role is managing conversation between the different agents to achieve this goal.

Below is a description of how AutoLabs works.
The chemicals are added to an array of vials. The dimensions of the array depend on the size of the vials used:
- 8x12 array = 1.2 mL vials
- 6x8 array = 2 mL vials
- 4x6 array = 4 or 8 mL vials
- 2x4 array = 20 mL vials
USE ONLY THESE PLATE SIZES. 

The rows are indexed with letters, A, B, C, D,.. etc. The columns are indexed with numbers, 1,2,3,4,.. etc.
Based on the description of the experiment and the user responses you should determine which size of vials are needed and based on the size of the vials determine the array size.

Here are the capabilities of the robot:
    - Add Chemical X (unit) to vials in Plate N { }
    - Set HeatingTemp in vials in Plate N { }
    - Set Cap vials in Plate N { }
    - Set Uncap vials in Plate N { }
    - Set Delay to X min in vials in Plate N { }
    - Set StirRate to X rpm in vials in Plate N { }
    - Set VortexRate in vials in Plate N { }
    - Set VialTimers in vials in Plate N { }

HeatingTemp, Cap, Uncap,  Delay, StirRate, VortexRate, VialTimers are processing actions. When generating the processing steps you must use exactly these 
terms when a processing action is requried. For example, don't use Heating. You should use HeatingTemp.

Each of these steps can be completed multiple times and in different orders depending on the needs of the experiment. Use your experimental chemistry knowledge to infer the correct step ordering and  parameters based on the information provided by the user. You should use your chemical knowledge to infer information such as the volume of vials needed, if the experiement should utilize multiple plates, the correct dispensing methods for chemicals, the amounts of chemicals to be added based on user descriptions, when and how long to stir, whether the solutions need to be heated, etc. 

YOU MUST USE THE PLATE NAME IN EACH STEP. IS DOES NOT MATTER WHAT KIND OF STEP IT IS. 

If multiple plates are required, be sure to use consitent plate names (i.e. Plate 1, Plate 2, Plate 3, etc.) and to specify the plate name in each step that involves a vial. The vials pertain to the plate that they are in. For example, A1 in plate 1 is different from A1 in plate 2. If a user specifies a plate, the robot should use the same plate for all subsequent steps unless otherwise specified. Still be sure to name the plate in each step. Additionally, if the user does not specify a plate for a step, the robot should ask which plates the step needs to apply to.
In some experiments, the user may need to transfer chemicals from one plate to another. In these steps, ensure each step is clear about the source and destination plate, and should only include maximum 2 plates per step. If more than 1 plate transfer is needed, use multiple steps. Additionally, you should always start this step with the word Transfer.

The user may need to specify many different chemicals depending on the experiment. If the step involves adding a chemical to vials, this is the format to utilize.

{ } is a dictionary of the format {vial_index1: value1, vial_index2: value2}, where the value could be amount of chemical added to a vial, heating temperature, etc. Each experiment step involves making changes to the vials. If needed, you can ask the neccessary input from the user regarding what changes happen to each vial. 
You must get this information in the following format,
    - <step> Description of the step. { } </step>.

{ } should not be a nested dictionary. Its' items should striclty be a key:value pair, where the key should be the vial index named as A1, A2, etc. The units of the values are as follows,

    - if the chemical is a solid, use mg.
    - if the chemical is a liquid, use uL.
    - Cap is 1 and Uncap is 0.
    - HeatingTemp is in celcius.

Do not include the units in the dictionary.

A few examples are below.
    - <step> Add chemical_name (unit) to vials in Plate 1. {A1: .1, A2:.3, D1:.5} </step>
    - <step> Set HeatingTemp to to 25 degC in Plate 1. {A1: 25, A2:25, D1:25} </step>
    - <step> Set Cap vials in Plate 1. {A1: 1, A2:1, D1:0} </step>
The dictionary SHOULD NOT contain ambiguous characters like "..." . The dictionary SHOULD contain key:value pairs.

You can also infer the amounts needed for each vials based on the user description. For example, if the user wants to synthesize a range of concentrations, use your chemical knowledge to infer the amounts of chemicals and solvents that need to be added. When performing calculations of amounts please be mindful of the specified concentration of the starting solutions. Units must be specified in mg for solids and ul for liquids even if the user provides other units in the input.

For steps that involve plate to plate transfers (i.e. moving a chemical from one plate to another), the format of the dictionary should be as follows:
{ } is a dictionary of the format {vial_index1_source_plate:[vial_index1_destination_plate, ammount], vial_index2_source_plate: [vial_index2_destination_plate, ammount]}, where the key is the plate being taken from, and the value is a list containing the plate being added to and the ammount that needs to be added. YOU MUST INCUDE THE UNIT IN THE AMMOUNT. If needed, you can ask the neccessary input from the user regarding which vials the chemical is being taken from and moved to.
There are two cases of plate to plate transfer - uniform represents that the ammount take from the source will be the same for all vials involved in the step, discrete represents that the ammount taken from the source may be different for each vial involved in the step. You will need to specify which type when you generate the step. If there are different amounts needing to be transfered, please assume discrete. If you do not know which type to use, ask the user for clarification. 

A few examples are below.
    - <step> Discrete transfer from plate 1 to plate 2. {A1:[a1, 5ul], A2:[a2, 5ul], A3:[a3, 10ul]} </step>
    - <step> Uniform transfer from plate 1 to plate 2. {A1:[a1, 5ul], A2:[a2, 5ul], A3:[a3, 5ul]} </step>

    
A variant of plate to plate transfers is when the user wants transfer from plate 1 to plate 2, wait for different time intervals and then tranfer back from plate 2 to plate 1.
The steps corresponding to this case looks like,
    - <step> Discrete transfer from plate 1 to plate 2. {A1:[a1, 5ul], A2:[a2, 5ul], A3:[a3, 10ul]} </step>
    - <step> Set VialTimers in Plate 1 {A1:10, A2:15, A3:20} </step>
    - <step> Uniform transfer from plate 2 to plate 1. {A1:[a1, 5ul], A2:[a2, 5ul], A3:[a3, 5ul]} </step>
This situation most arises when doing kinetic studies where each vial within a plate, usually with the same starting chemical composition, will be heated for a different amount of time. 

For efficiency, please use as few steps as possible for adding each input chemical. Add a given chemical to all vials where it will be needed in the same step, unless there is an experimental reason to do it in multiple steps.

Vial size should be determined based on the target working volumes of the experiment.  
- There are seven standard vials sizes (associated array listed as rows x columns).
- 1mL (8x12), 1.2mL (8x12), 2mL (6x8), 4mL (4x6), 8mL (4x6), 20mL (2x4), and 125mL (1x2)
- The working volume of prepared samples should be between 10 - 80% of the total vial volume.
- 1mL, 1.2mL, and 125mL vials cannot be capped or uncapped automatically.

In chemical addition steps, you MUST use proper chemical names.

When calculating the solvent volume, you SHOULD take into account the volumes of all the other chemicals in the solution.

When a vial requires both solids and liquids to be added, the Solid addition must be first.  The only exception to adding liquids before solids, is if the liquid/solution is aqueous.

Vials containing volatile liquids (usually non-aqueous samples), should be kept capped, when possible, to ensure the liquids do not evaporate during preparation.

HeatingTemp steps must be between 25 - 180 deg C. Vials should be capped before heating.

StirRate steps are limited to 700 rpm.

VortexRate steps are limited to 1000 rpm.

If a solvent is needed, make sure to specify the solvent in the step. For example, you should not just say Add solvent to vials in plate 1. You should say Add water solvent to vials in plate 1.

If an experiment uses vortex rate or stir, make sure to zero it after the delay. Vortexing is preferred to stirring unless otherwise specified by the user.

Using the neccessary capabilities from the above list, generate the steps one at a time. Ask as many clarification questions as needed but try to use your chemistry knowledge to infer the neccessary steps from the general descriptions provided by the user.

Using your chemistry knowledge, make sure to think carefully about whether the vials need to heated, stirred, capped, or rested during the experimental procedures.

Each step should utilize Set, Add, or Transfer to describe what is occuring to the plate and vials. 

If a modifier A is dispensed as a solution of n% A in the solvent B, this means that the modifier is a pre-mixed solution of n% (v/v) A dissolved in solvent B. Do not use pure A in the calcualtion. All A additions come from this pre-diluted stock.

Once clear about all the steps, you should print all the steps using <step> tags. Enclose these final steps using <final-steps> tag. Make sure there are no newlines between the <final-steps> and <step> tags. Do not use any special characters in the steps. Please list all of the <steps></steps> within one set of <final-steps> tags.

Please get confirmation from the user that your reasoning is correct before generating the <final-steps> tags.

When there is a step for set HeatingTemp and set VialTimers in the description, there will be two transfer steps. The first transfer step will need to include "StartVialTimers" in the sep description and the second will need to include "WaitVialTimers" in the step description. Also include "MoveVial" in the description. Set the VialTimers before either of the transfer steps.
For example:
    - <step> Set VialTimers in Plate 1 {A1:10, A2:15, A3:20} </step>
    - <step> Set HeatingTemp to to 25 degC in Plate 2. {A1: 25, A2:25, D1:25} </step>
    - <step> Uniform transfer from plate 1 to plate 2. (MoveVial, StartVialTimer) {A1:[a1, 5ul], A2:[a2, 5ul], A3:[a3, 5ul]} </step>
    - <step> Uniform transfer from plate 2 to plate 1. (MoveVial, WaitVialTimer) {A1:[a1, 5ul], A2:[a2, 5ul], A3:[a3, 5ul]} </step>


When you receive the experiment description, following the following steps may make it easy.

0: 'Refine and correct experimental steps according to user instructions. Confirm with the user the agent's understanding of the experiment and source chemicals',
1: 'For each reaction or mixture, perform calculations.',
2: 'Determine the vial organization and assign reactions to specific vials.',
3: 'Determine additional processing steps.',
4: 'Confirm with user before generating final steps.'

If you are not absolutely sure about the steps and the calculations, DO NOT jump straight to decide the final steps. GO STEP BY STEP.

The role of each agent is listed below.
| Agent | Role |
| ----- | ---- |
| Undesrand_And_Refine_Experiment | 'Refine and correct experimental steps according to user instructions. Confirm with the user the agent's understanding of the experiment and source chemicals' |
| Calculate_Chemical_Amounts_For_Reactions | 'For each reaction or mixture, perform calculations.',|
| Determine_Vial_Organization | 'Determine the vial organization and assign reactions to specific vials.',|
| Determine_Processing_Steps | 'Determine additional processing steps.',|
| Generate_Final_Steps | Gather, organize and generate final steps using <final-steps> tags|

A request on math expression rendering: please enclose math expressions with $ for nice streamit rendering.
\end{lstlisting}

\clearpage
\section{Hardware Options}
This section details the available options and tags that can be used to refine the chemical addition steps, providing a structured approach to customizing the experimental procedures. It includes specific tags for both solid and liquid chemical additions, as well as dispensing methods and associated parameters.

\begin{enumerate}
    \item Chemical Addition for Solids:
    \begin{enumerate}
        \item  Core Tag:
        \begin{itemize}
            \item Powder: Indicates the addition involves a powdered chemical.
        \end{itemize}

        \item Optional Tags
        \begin{itemize}
            \item Plate: Enables efficient addition to an entire plate.
            \item Notify: Sends alerts to operators when intervention or confirmation is required.
        \end{itemize}

    \end{enumerate}

    \item Chemical Addition for Liquids:

    \begin{itemize}
        \item Core Dispensing Methods:
      - SyringePump: Ideal for dispensing water or miscible liquids.
      - PDT: Recommended for dispensing immiscible liquids.
        \item Tags for SyringePump Additions:
        \begin{itemize}
            \item Backsolvent: Manages residual solvent for improvements in dispensing.
            \item ExtSingleTip: Specifies dispensing with a single extended tip.
      \item 4Tip: Allows for multi-tip dispensing to enhance speed.
      \item LookAhead: Predictive optimization for smoother operations.
      \item SourceTracking and DestinationTracking: Tracks where chemicals are sourced from and where they are dispensed.
      \item Hover: Keeps the tip above liquid height during dispensing.
      \item StartVialTimer and WaitVialTimer: Manages timed operations linked to vial use.
      \item Notify: Alerts operators as needed during operations.
        \end{itemize}

        \item Tags for PDT Additions:
        \begin{itemize}
            \item Tip Size: Specifies the required tip type (e.g., 10mLTip, 1000uLTip).
            \item Same optional tags as listed above (Backsolvent, Hover, Notify, etc.) can be applied.
        \end{itemize}

    \end{itemize}

\end{enumerate}

\clearpage
\section{Experiment Details}
\subsection{Experiment 1}

Description: Prepare a set of eight calibration samples for naphthalene in methanol. Total volume of each sample is 10 mL. Samples should have 5,10,15,20,25,30, 35 and 50 mg of naphthalene with the remainder being methanol. After preparation, the vials should be capped and then mixed via vortexing for 10 minutes. No heating required.

\subsubsection*{Ground Truth Steps}
\begin{enumerate}[itemsep=0pt]
    \item Add naphthalene (mg) to Plate 1
    \item Add methanol (ul) to Plate 1
    \item Set Cap in Plate 1
    \item Set VortexRate to 500 in Plate 1
    \item Set Delay to 10 min in Plate 1
    \item Set VortexRate to 0 in Plate 1
\end{enumerate}

\subsubsection*{Ground Truth Chemical Amounts}

\begin{table*}[!htb]
    \centering
    \scriptsize
    \begin{tabular}{lrr}
    \toprule
    Vial & naphthalene & methanol \\
    \midrule
    A1 & 5 & 9995.61 \\
    A2 & 10 & 9991.23 \\
    A3 & 15 & 9986.84 \\
    A4 & 20 & 9982.46 \\
    B1 & 25 & 9978.07 \\
    B2 & 30 & 9973.68 \\
    B3 & 35 & 9969.30 \\
    B4 & 50 & 9956.14 \\
    \bottomrule
    \end{tabular}
    \caption{Ground truth chemical quantities for Experiment 1. Units: naphthalene: mg, methanol: $\mu l$}
    \label{tab:placeholder}
\end{table*}

\subsubsection*{Calculations}
\begin{itemize}
    \item m\_naphthalene (g) = (target mass in mg) ÷ 1000
    \item V\_naphthalene (mL) = m\_naphthalene (g) $\div$ $\rho$\_naphthalene (1.14 g mL–1)
    \item V\_methanol (mL) = 10.000 mL – V\_naphthalene (mL)
    \item C\_naphthalene (mg mL–1) = (mass in mg) ÷ 10.000 mL
\end{itemize}

\subsubsection*{Example calculation for 5 mg of naphthalene:}
m\_naph = 5 mg $\div$ 1 000 = 0.005 g \\
V\_naph = 0.005 g $\div$ 1.14 g mL–1 = 0.00439 mL \\
V\_MeOH = 10.000 mL - 0.00439 mL = 9.99561 mL \\

\subsection{Experiment 2}
Description: Prepare a set of electrolyte solutions, each consisting of a salt, solvent, and modifier. I
have three salts: lithium perchlorate, lithium tetrafluoroborate and lithium hexafluorophosphate. I want to prepare
6 compositions with salt 1 and 2, loading the vials at 20 mg each, and 12 compositions with salt 3, loading half the
vials with 20 mg and half with 50 mg. The solvent is propylene carbonate and the modifier is ethylene carbonate,
which will be dispensed as a solution of 1\% ethylene carbonate in propylene carbonate. For each salt loading I want
500 uL to be the final volume, and the modifier concentration should be varied from 0\%, 0.2\%, 0.4\%, 0.6\%, 0.8\%,
and 1.0\%. After preparing these solutions, they should be heated to 40 deg for 30 minutes with stirring to ensure
homogeneity. 

\subsubsection*{Ground Truth Steps}

\begin{enumerate}[itemsep=0pt]
    \item Add lithium perchlorate (mg) to vials in Plate 1
      \item Add lithium tetrafluoroorate (mg) to vials in Plate 1
      \item Add lithium hexafluorophosphate (mg) to vials in Plate 1
      \item Add 1\% ethylene carbonate (ul) to vials in Plate 1
      \item Add propylene carbonate (ul) to vials in Plate 1
      \item Set Cap to vials in Plate 1
      \item Set StirRate to 700 rpm in Plate 1
      \item Set HeatingTemp to 40 C in Plate 1
      \item Set Delay to 30 min in Plate 1
      \item Set HeatingTemp to 25 C in Plate 1
      \item Set StirRate to 0 rpm in Plate 1
\end{enumerate}

\subsubsection*{Ground Truth Chemical Amounts}
\begin{table*}[!htb]
    \centering
    \scriptsize
    \begin{tabular}{lrrrrr}
    \toprule
    Vial & LP & PC & 1\% EC in PC & LT & LH \\
    \midrule
    A1 & 20 & 500 & 0 & 0 & 0 \\
    A2 & 20 & 400 & 100 & 0 & 0 \\
    A3 & 20 & 300 & 200 & 0 & 0 \\
    A4 & 20 & 200 & 300 & 0 & 0 \\
    A5 & 20 & 100 & 400 & 0 & 0 \\
    A6 & 20 & 0 & 500 & 0 & 0 \\
    B1 & 0 & 500 & 0 & 20 & 0 \\
    B2 & 0 & 400 & 100 & 20 & 0 \\
    B3 & 0 & 300 & 200 & 20 & 0 \\
    B4 & 0 & 200 & 300 & 20 & 0 \\
    B5 & 0 & 100 & 400 & 20 & 0 \\
    B6 & 0 & 0 & 500 & 20 & 0 \\
    C1 & 0 & 500 & 0 & 0 & 20 \\
    C2 & 0 & 400 & 100 & 0 & 20 \\
    C3 & 0 & 300 & 200 & 0 & 20 \\
    C4 & 0 & 200 & 300 & 0 & 20 \\
    C5 & 0 & 100 & 400 & 0 & 20 \\
    C6 & 0 & 0 & 500 & 0 & 20 \\
    D1 & 0 & 500 & 0 & 0 & 50 \\
    D2 & 0 & 400 & 100 & 0 & 50 \\
    D3 & 0 & 300 & 200 & 0 & 50 \\
    D4 & 0 & 200 & 300 & 0 & 50 \\
    D5 & 0 & 100 & 400 & 0 & 50 \\
    D6 & 0 & 0 & 500 & 0 & 50 \\
    \bottomrule
    \end{tabular}
    \caption{Ground truth chemical quantities for Experiment 2. Units: naphthalene: mg, methanol: $\mu l$}
    \label{tab:placeholder}
\end{table*}

\subsubsection*{Calculations}
EC = ethylene carbonate \\
PC = propylene carbonate  \\

\noindent For a final vial volume $V_{total}$ = 500 $\mu$L and desired modifier concentration $C_{mod}$ ($\%~v/v$ EC in the final solution), we use a 1 \%~EC/PC stock solution (i.e. 1 $\mu$L EC + 99 $\mu$L PC).\\

\noindent Volume of 1\% EC/PC stock to add, $V_{stock}$:
\begin{equation}
    V_{mod}= \left( \frac{C_{mod} }{1\% } \right) \times V_{total}
\end{equation}

\noindent Volume of neat PC to add, $V_{PC}$:
\begin{equation}
    V_{PC} = V_{total} - V_{mod}
\end{equation}

\subsubsection*{Example for $C_{mod}$ = 0.4 \%:}

$V_{mod}$ = (0.4 \% $\div$ 1 \%) $\times$ 500 $\mu$ L = 0.4 $\times$ 500 $\mu$L = 200 $\mu$L\\
$V_{PC}$ = 500 $\mu$L - 200 $\mu$L = 300 $\mu$L \\

\subsection{Experiment 3}
Description: Perform a set of imine synthesis experiments using aqueous ammonia as the nitrogen source and solvent for the reaction.  Each reaction will be done in duplicate.  Each vial will contain two reactants: R1 and R2.  R1 is benzaldehyde at 0.5 mmol loading.  R2 will be one of 8 compounds, loaded at 0.75 mmol.  R2 chemicals are:  1-bromobutane, 1-iodobutane, 1-chlorobutane, 3-bromopropene, benzyl bromide, 3-bromobut-1-ene, 3-bromobut-2-ene, and 2-bromoethyl cyanide.  We want to see how the amount of ammonia affects the overall product yields, so the amount of water and 28\% aqueous ammonia solution should be calculated as to achieve ammonia loadings of 3M, 9M and 12M.  The total solution volume will be 1mL.  The samples will be headed at 60 degC overnight.

\subsubsection*{Ground Truth Steps}

\begin{enumerate}[itemsep=0pt]
    \item Add water (ul) to vials in Plate 1
    \item Add aqueous ammonia (ul) to vials in Plate 1
    \item Add benzaldehyde (ul) to vials in Plate 1
    \item Add 1-bromobutane (ul) to vials in Plate 1
    \item Add 1-iodobutane (ul) to vials in Plate 1
    \item Add 1-chlorobutane (ul) to vials in Plate 1
    \item Add 3-bromopropene (ul) to vials in Plate 1
    \item Add benzyl bromide (ul) to vials in Plate 1
    \item Add 3-bromobut-1-ene (ul) to vials in Plate 1
    \item Add 3-bromobut-2-ene (ul) to vials in Plate 1
    \item Add 2-bromoethyl cyanide (ul) to vials in Plate 1
    \item Set Cap for vials in Plate 1
    \item Set StirRate to 700 rpm in Plate 1
    \item Set HeatingTemp to 60 C in Plate 1
    \item Set Delay to 480 min in Plate 1
    \item Set HeatingTemp to 25 C in Plate 1
    \item Set StirRate to 0 rpm in Plate 1
\end{enumerate}

\begin{table*}[!htb]
    \centering
    \scriptsize
\begin{tabular}{lrrrrrrrrrrr}
\toprule
Vial & water & NH3 & benzald & 1-bromobu & 1-iodobu & 1-chlorobu & 3-bromoprop & benzyl\_b & 3-bromobu1 & 3-bromobu2 & 2-bromoeth \\
\midrule
A1 & 664.45 & 203.67 & 50.83 & 81.06 & 0.00 & 0.00 & 0.00 & 0.00 & 0.00 & 0.00 & 0.00 \\
A2 & 660.16 & 203.67 & 50.83 & 0.00 & 85.35 & 0.00 & 0.00 & 0.00 & 0.00 & 0.00 & 0.00 \\
A3 & 667.50 & 203.67 & 50.83 & 0.00 & 0.00 & 78.01 & 0.00 & 0.00 & 0.00 & 0.00 & 0.00 \\
A4 & 681.05 & 203.67 & 50.83 & 0.00 & 0.00 & 0.00 & 64.46 & 0.00 & 0.00 & 0.00 & 0.00 \\
A5 & 658.65 & 203.67 & 50.83 & 0.00 & 0.00 & 0.00 & 0.00 & 86.86 & 0.00 & 0.00 & 0.00 \\
A6 & 668.81 & 203.67 & 50.83 & 0.00 & 0.00 & 0.00 & 0.00 & 0.00 & 76.70 & 0.00 & 0.00 \\
A7 & 669.94 & 203.67 & 50.83 & 0.00 & 0.00 & 0.00 & 0.00 & 0.00 & 0.00 & 75.57 & 0.00 \\
A8 & 674.11 & 203.67 & 50.83 & 0.00 & 0.00 & 0.00 & 0.00 & 0.00 & 0.00 & 0.00 & 71.39 \\
B1 & 257.15 & 610.97 & 50.83 & 81.06 & 0.00 & 0.00 & 0.00 & 0.00 & 0.00 & 0.00 & 0.00 \\
B2 & 252.86 & 610.97 & 50.83 & 0.00 & 85.35 & 0.00 & 0.00 & 0.00 & 0.00 & 0.00 & 0.00 \\
B3 & 260.20 & 610.97 & 50.83 & 0.00 & 0.00 & 78.01 & 0.00 & 0.00 & 0.00 & 0.00 & 0.00 \\
B4 & 273.75 & 610.97 & 50.83 & 0.00 & 0.00 & 0.00 & 64.46 & 0.00 & 0.00 & 0.00 & 0.00 \\
B5 & 251.35 & 610.97 & 50.83 & 0.00 & 0.00 & 0.00 & 0.00 & 86.86 & 0.00 & 0.00 & 0.00 \\
B6 & 261.51 & 610.97 & 50.83 & 0.00 & 0.00 & 0.00 & 0.00 & 0.00 & 76.70 & 0.00 & 0.00 \\
B7 & 262.64 & 610.97 & 50.83 & 0.00 & 0.00 & 0.00 & 0.00 & 0.00 & 0.00 & 75.57 & 0.00 \\
B8 & 266.81 & 610.97 & 50.83 & 0.00 & 0.00 & 0.00 & 0.00 & 0.00 & 0.00 & 0.00 & 71.39 \\
C1 & 53.50 & 814.62 & 50.83 & 81.06 & 0.00 & 0.00 & 0.00 & 0.00 & 0.00 & 0.00 & 0.00 \\
C2 & 49.20 & 814.62 & 50.83 & 0.00 & 85.35 & 0.00 & 0.00 & 0.00 & 0.00 & 0.00 & 0.00 \\
C3 & 56.55 & 814.62 & 50.83 & 0.00 & 0.00 & 78.01 & 0.00 & 0.00 & 0.00 & 0.00 & 0.00 \\
C4 & 70.10 & 814.62 & 50.83 & 0.00 & 0.00 & 0.00 & 64.46 & 0.00 & 0.00 & 0.00 & 0.00 \\
C5 & 47.69 & 814.62 & 50.83 & 0.00 & 0.00 & 0.00 & 0.00 & 86.86 & 0.00 & 0.00 & 0.00 \\
C6 & 57.86 & 814.62 & 50.83 & 0.00 & 0.00 & 0.00 & 0.00 & 0.00 & 76.70 & 0.00 & 0.00 \\
C7 & 58.99 & 814.62 & 50.83 & 0.00 & 0.00 & 0.00 & 0.00 & 0.00 & 0.00 & 75.57 & 0.00 \\
C8 & 63.16 & 814.62 & 50.83 & 0.00 & 0.00 & 0.00 & 0.00 & 0.00 & 0.00 & 0.00 & 71.39 \\
D1 & 664.45 & 203.67 & 50.83 & 81.06 & 0.00 & 0.00 & 0.00 & 0.00 & 0.00 & 0.00 & 0.00 \\
D2 & 660.16 & 203.67 & 50.83 & 0.00 & 85.35 & 0.00 & 0.00 & 0.00 & 0.00 & 0.00 & 0.00 \\
D3 & 667.50 & 203.67 & 50.83 & 0.00 & 0.00 & 78.01 & 0.00 & 0.00 & 0.00 & 0.00 & 0.00 \\
D4 & 681.05 & 203.67 & 50.83 & 0.00 & 0.00 & 0.00 & 64.46 & 0.00 & 0.00 & 0.00 & 0.00 \\
D5 & 658.65 & 203.67 & 50.83 & 0.00 & 0.00 & 0.00 & 0.00 & 86.86 & 0.00 & 0.00 & 0.00 \\
D6 & 668.81 & 203.67 & 50.83 & 0.00 & 0.00 & 0.00 & 0.00 & 0.00 & 76.70 & 0.00 & 0.00 \\
D7 & 669.94 & 203.67 & 50.83 & 0.00 & 0.00 & 0.00 & 0.00 & 0.00 & 0.00 & 75.57 & 0.00 \\
D8 & 674.11 & 203.67 & 50.83 & 0.00 & 0.00 & 0.00 & 0.00 & 0.00 & 0.00 & 0.00 & 71.39 \\
E1 & 257.15 & 610.97 & 50.83 & 81.06 & 0.00 & 0.00 & 0.00 & 0.00 & 0.00 & 0.00 & 0.00 \\
E2 & 252.86 & 610.97 & 50.83 & 0.00 & 85.35 & 0.00 & 0.00 & 0.00 & 0.00 & 0.00 & 0.00 \\
E3 & 260.20 & 610.97 & 50.83 & 0.00 & 0.00 & 78.01 & 0.00 & 0.00 & 0.00 & 0.00 & 0.00 \\
E4 & 273.75 & 610.97 & 50.83 & 0.00 & 0.00 & 0.00 & 64.46 & 0.00 & 0.00 & 0.00 & 0.00 \\
E5 & 251.35 & 610.97 & 50.83 & 0.00 & 0.00 & 0.00 & 0.00 & 86.86 & 0.00 & 0.00 & 0.00 \\
E6 & 261.51 & 610.97 & 50.83 & 0.00 & 0.00 & 0.00 & 0.00 & 0.00 & 76.70 & 0.00 & 0.00 \\
E7 & 262.64 & 610.97 & 50.83 & 0.00 & 0.00 & 0.00 & 0.00 & 0.00 & 0.00 & 75.57 & 0.00 \\
E8 & 266.81 & 610.97 & 50.83 & 0.00 & 0.00 & 0.00 & 0.00 & 0.00 & 0.00 & 0.00 & 71.39 \\
F1 & 53.50 & 814.62 & 50.83 & 81.06 & 0.00 & 0.00 & 0.00 & 0.00 & 0.00 & 0.00 & 0.00 \\
F2 & 49.20 & 814.62 & 50.83 & 0.00 & 85.35 & 0.00 & 0.00 & 0.00 & 0.00 & 0.00 & 0.00 \\
F3 & 56.55 & 814.62 & 50.83 & 0.00 & 0.00 & 78.01 & 0.00 & 0.00 & 0.00 & 0.00 & 0.00 \\
F4 & 70.10 & 814.62 & 50.83 & 0.00 & 0.00 & 0.00 & 64.46 & 0.00 & 0.00 & 0.00 & 0.00 \\
F5 & 47.69 & 814.62 & 50.83 & 0.00 & 0.00 & 0.00 & 0.00 & 86.86 & 0.00 & 0.00 & 0.00 \\
F6 & 57.86 & 814.62 & 50.83 & 0.00 & 0.00 & 0.00 & 0.00 & 0.00 & 76.70 & 0.00 & 0.00 \\
F7 & 58.99 & 814.62 & 50.83 & 0.00 & 0.00 & 0.00 & 0.00 & 0.00 & 0.00 & 75.57 & 0.00 \\
F8 & 63.16 & 814.62 & 50.83 & 0.00 & 0.00 & 0.00 & 0.00 & 0.00 & 0.00 & 0.00 & 71.39 \\
\bottomrule
\end{tabular}
    \caption{Ground truth chemical quantities for Experiment 3.}
    \label{tab:placeholder}
\end{table*}

\subsubsection*{Calculations}
R1 = benzaldehyde

\begin{equation*}
    Mass~of~R_{1}=0.5mmol×(106.12g/mol)=0.05306g
\end{equation*}
\begin{equation*}
    Volume~of~R_{1}= \frac{Mass~of~R_{1}}{Density} = \frac{0.05306g}{1.045g/mL} \approx 50.8  \mu L
\end{equation*}

Calculation steps are similar for $R_{2}$.

Calculation of $NH_{3}$ volume.

\begin{itemize}
    \item[] Molecular weight of 28\% NH3:
        \begin{itemize}
            \item[] MW = 17.031 g/mol
        \end{itemize}

    \item[] Molecular density:
        \begin{itemize}
            \item[] D = 0.896 g/ml
        \end{itemize}

    \item[] Calculate the mass of NH3 in 1 L of solution:
        \begin{itemize}
            \item[] mass = 1000 * D = 896 g
        \end{itemize}

    \item[] Mass of 28\%:

        \begin{itemize}
            \item[] mass\_n\_percent = 28 * mass/100 = 28 * 896/ 100 = 250.88 g
        \end{itemize}

    \item[] Calculate the number of moles

    \begin{itemize}
        \item[] moles = mass\_n\_percent/MW = 250.88/ 17.031 = 14.73
    \end{itemize}

    \item[]  molarity = 14.73/ 1 L = 14.73
 
    \item[] Use the dilution equation to find V1 for 3M case.
    \item[] Dilution equation => C1V1 = C2V2
 
     \item[]  C1 = 14.73, V1=?, C2=3, V2 = 0.001 L
            \begin{itemize}
                 \item[] V1 = C2*V2/C1
                 \item[] = 3 * 0.001 / 14.73
                 \item[] = 203 uL
            \end{itemize}
\end{itemize}

\subsection{Experiment 4}
Description: Perform esterification reactions between an acid and an alcohol using the liquid catalyst sulfuric acid.  I would like to use the following alcohols: methanol, ethanol, propanol, and glycerol.  I have three acids:  acetic acid, propanoic acid, and benzoic acid.  Acetic acid should be combined with all four alcohols, but benzoic acid and propanoic acid should only be combined with methanol and ethanol.  I would like the molar ratio between the alcohol and acid to be examined at 0.5, 1.0 and 2.0, with the total Molarity of the acid and alcohol to be kept constant at 4 M.  The liquid catalyst should be set to 0.025 M, added as a 0.5M solution of sulfuric acid in water.  The remaining 2ml volume should be water.  After the solutions are prepared, they should be heated to 80 deg for 30 mins before cooling completely to 25C.  Once cooled, a portion of the sample can be diluted by a DF of 10 and transferred to HPLC vials (1mL total volume).  The HPLC samples should be vortexed for 20 minutes after being prepared.

\subsubsection*{Ground Truth Steps}
\begin{enumerate}[itemsep=0pt]
    \item Add benzoic acid (mg) to vials in Plate 1
    \item Add water (ul) to vials in Plate 1
    \item Add acetic acid (ul) to vials in Plate 1
    \item Add propanoic acid (ul) to vials in Plate 1
    \item Add methanol (ul) to vials in Plate 1
    \item Add ethanol (ul) to vials in Plate 1
    \item Add propanol (ul) to vials in Plate 1
    \item Add glycerol (ul) to vials in Plate 1
    \item Add sulfuric acid (ul) to vials in Plate 1
    \item Set Cap vials in Plate 1
    \item Set StirRate in Plate 1
    \item Set HeatingTemp in Plate 1
    \item Set Delay in Plate 1
    \item Set HeatingTemp in Plate 1
    \item Set StirRate in Plate 1
    \item Add water (ul) to vials in Plate 2
    \item Discrete transfer from Plate 1 to Plate 2
    \item Set Cap vials in Plate 2
    \item Set VortexRate in Plate 2
    \item Set Delay in Plate 2
    \item Set VortexRate in Plate 2
\end{enumerate}

\begin{table}[!htb]
    \scriptsize
\begin{tabular}{lrrrrrrrrr}
\toprule
Vial & water & acetic acid & methanol & sulfuric & propanoic & ethanol & propanol & benzoic & glycerol \\
\midrule
A1 & 1531.32 & 152.65 & 216.03 & 100 & 0.00 & 0.00 & 0.00 & 0.00 & 0.00 \\
A2 & 1485.03 & 0.00 & 216.03 & 100 & 198.94 & 0.00 & 0.00 & 0.00 & 0.00 \\
A3 & 1509.00 & 228.98 & 162.02 & 100 & 0.00 & 0.00 & 0.00 & 0.00 & 0.00 \\
A4 & 1439.57 & 0.00 & 162.02 & 100 & 298.41 & 0.00 & 0.00 & 0.00 & 0.00 \\
A5 & 1486.68 & 305.31 & 108.02 & 100 & 0.00 & 0.00 & 0.00 & 0.00 & 0.00 \\
A6 & 1394.11 & 0.00 & 108.02 & 100 & 397.88 & 0.00 & 0.00 & 0.00 & 0.00 \\
B1 & 1435.93 & 152.65 & 0.00 & 100 & 0.00 & 311.42 & 0.00 & 0.00 & 0.00 \\
B2 & 1389.65 & 0.00 & 0.00 & 100 & 198.94 & 311.42 & 0.00 & 0.00 & 0.00 \\
B3 & 1437.46 & 228.98 & 0.00 & 100 & 0.00 & 233.56 & 0.00 & 0.00 & 0.00 \\
B4 & 1368.03 & 0.00 & 0.00 & 100 & 298.41 & 233.56 & 0.00 & 0.00 & 0.00 \\
B5 & 1438.99 & 305.31 & 0.00 & 100 & 0.00 & 155.71 & 0.00 & 0.00 & 0.00 \\
B6 & 1346.41 & 0.00 & 0.00 & 100 & 397.88 & 155.71 & 0.00 & 0.00 & 0.00 \\
C1 & 1347.58 & 152.65 & 0.00 & 100 & 0.00 & 0.00 & 399.77 & 0.00 & 0.00 \\
C2 & 1683.97 & 0.00 & 216.03 & 100 & 0.00 & 0.00 & 0.00 & 325.65 & 0.00 \\
C3 & 1371.19 & 228.98 & 0.00 & 100 & 0.00 & 0.00 & 299.83 & 0.00 & 0.00 \\
C4 & 1737.98 & 0.00 & 162.02 & 100 & 0.00 & 0.00 & 0.00 & 488.48 & 0.00 \\
C5 & 1394.81 & 305.31 & 0.00 & 100 & 0.00 & 0.00 & 199.88 & 0.00 & 0.00 \\
C6 & 1791.98 & 0.00 & 108.02 & 100 & 0.00 & 0.00 & 0.00 & 651.31 & 0.00 \\
D1 & 1357.86 & 152.65 & 0.00 & 100 & 0.00 & 0.00 & 0.00 & 0.00 & 389.49 \\
D2 & 1588.58 & 0.00 & 0.00 & 100 & 0.00 & 311.42 & 0.00 & 325.65 & 0.00 \\
D3 & 1378.90 & 228.98 & 0.00 & 100 & 0.00 & 0.00 & 0.00 & 0.00 & 292.12 \\
D4 & 1666.44 & 0.00 & 0.00 & 100 & 0.00 & 233.56 & 0.00 & 488.48 & 0.00 \\
D5 & 1399.95 & 305.31 & 0.00 & 100 & 0.00 & 0.00 & 0.00 & 0.00 & 194.74 \\
D6 & 1744.29 & 0.00 & 0.00 & 100 & 0.00 & 155.71 & 0.00 & 651.31 & 0.00 \\
\bottomrule
\end{tabular}
    \caption{Ground truth chemical quantities for Experiment 4.}
    \label{tab:placeholder}
\end{table}

\subsubsection*{Calculations}

\begin{equation}
    R = \frac{[Alocohol]}{[Acid]}
\end{equation}

\begin{equation}
    [Alocohol] + [Acid] = 4M
\end{equation}

\begin{equation}
    [Acid] = \frac{4.0}{R+1}
\end{equation}

\begin{equation}
    [Alcohol] = R \times \frac{4.0}{R+1}
\end{equation}

\noindent Example calculations for acetic acid ans methanol for molar ratio, R =0.5

\begin{equation*}
    [acetic] = \frac{4}{0.5+1} = 2.67M
\end{equation*}

\begin{equation*}
    [methanol] = 0.5 \times [acetic] = 1.33M
\end{equation*}

\begin{equation*}
    V_{acetic} = 2.67 M \times 0.002 L \frac{molar\_mass_{acetic}}{density_{acetic}}
\end{equation*}

\begin{equation*}
    V_{methanol} = 1.33 M \times 0.002 L \frac{molar\_mass_{methanol}}{density_{methanol}}
\end{equation*}

\subsection{Experiment 5}
Description: Perform time studies for the esterification reactions between acetic acid and 4 different alcohols (methanol, ethanol, propanol, and glycerol).  The molar ratio between acid and alcohol will be 1:1, with the total molarity of the acid and alcohol to be kept at a constant 4M. Sulfuric acid will act as the catalyst for this reaction.  We will target 0.025M in the solution, added as a 0.5M stock solution of sulfuric acid in water.  The remaining 2ml volume in each sample will be water.  Each mixture will be reacted at 80 degC at 6 time points (15, 30, 60, 90, 120, and 150 minutes).

\subsubsection*{Ground Truth Steps}
\begin{enumerate}[itemsep=0pt]
    \item Add acetic acid (ul) to vials in Plate 1
    \item Add Methanol (ul) to vials in Plate 1
    \item Add ethanol (ul) to vials in Plate 1
    \item Add propanol (ul) to vials in Plate 1
    \item Add glycerol (ul) to vials in Plate 1
    \item Add sulfuric acid (ul) to vials in Plate 1
    \item Add Water (ul) to vials in Plate 1
    \item Set Cap in vials in Plate 1
    \item Set HeatingTemp in vials in Plate 2
    \item Set VialTimers in vials in Plate 1
    \item Set StirRate in Plate 2
    \item Uniform transfer from Plate 1 to Plate 2
    \item Uniform transfer from Plate 2 to Plate 1
    \item Set StirRate in Plate 2
    \item Set HeatingTemp in vials in Plate 2
\end{enumerate}

\begin{table}[!htb]
    \scriptsize
\begin{tabular}{lrrrrrrr}
\toprule
Vial & water & acetic & methanol & sulfuric & ethanol & propanol & glycerol \\
\midrule
A1 & 1509.00 & 228.98 & 162.02 & 100 & 0.00 & 0.00 & 0.00 \\
A2 & 1509.00 & 228.98 & 162.02 & 100 & 0.00 & 0.00 & 0.00 \\
A3 & 1509.00 & 228.98 & 162.02 & 100 & 0.00 & 0.00 & 0.00 \\
A4 & 1509.00 & 228.98 & 162.02 & 100 & 0.00 & 0.00 & 0.00 \\
A5 & 1509.00 & 228.98 & 162.02 & 100 & 0.00 & 0.00 & 0.00 \\
A6 & 1509.00 & 228.98 & 162.02 & 100 & 0.00 & 0.00 & 0.00 \\
B1 & 1437.46 & 228.98 & 0.00 & 100 & 233.56 & 0.00 & 0.00 \\
B2 & 1437.46 & 228.98 & 0.00 & 100 & 233.56 & 0.00 & 0.00 \\
B3 & 1437.46 & 228.98 & 0.00 & 100 & 233.56 & 0.00 & 0.00 \\
B4 & 1437.46 & 228.98 & 0.00 & 100 & 233.56 & 0.00 & 0.00 \\
B5 & 1437.46 & 228.98 & 0.00 & 100 & 233.56 & 0.00 & 0.00 \\
B6 & 1437.46 & 228.98 & 0.00 & 100 & 233.56 & 0.00 & 0.00 \\
C1 & 1371.19 & 228.98 & 0.00 & 100 & 0.00 & 299.83 & 0.00 \\
C2 & 1371.19 & 228.98 & 0.00 & 100 & 0.00 & 299.83 & 0.00 \\
C3 & 1371.19 & 228.98 & 0.00 & 100 & 0.00 & 299.83 & 0.00 \\
C4 & 1371.19 & 228.98 & 0.00 & 100 & 0.00 & 299.83 & 0.00 \\
C5 & 1371.19 & 228.98 & 0.00 & 100 & 0.00 & 299.83 & 0.00 \\
C6 & 1371.19 & 228.98 & 0.00 & 100 & 0.00 & 299.83 & 0.00 \\
D1 & 1378.90 & 228.98 & 0.00 & 100 & 0.00 & 0.00 & 292.12 \\
D2 & 1378.90 & 228.98 & 0.00 & 100 & 0.00 & 0.00 & 292.12 \\
D3 & 1378.90 & 228.98 & 0.00 & 100 & 0.00 & 0.00 & 292.12 \\
D4 & 1378.90 & 228.98 & 0.00 & 100 & 0.00 & 0.00 & 292.12 \\
D5 & 1378.90 & 228.98 & 0.00 & 100 & 0.00 & 0.00 & 292.12 \\
D6 & 1378.90 & 228.98 & 0.00 & 100 & 0.00 & 0.00 & 292.12 \\
\bottomrule
\end{tabular}
    \caption{Ground truth chemical quantities for Experiment 5.}
    \label{tab:placeholder}
\end{table}

\clearpage
\section{Precision and Recall heatmaps corresponding to Parameter step generation}
This section presents heatmaps illustrating the precision and recall of the parameter step generation process. These heatmaps provide a visual representation of the performance of the system across different experiments, highlighting areas of strength and potential improvement.

\begin{figure}[htbp]
\centering
\includegraphics[width=1\linewidth, trim={0 4cm 0 5cm}, clip]{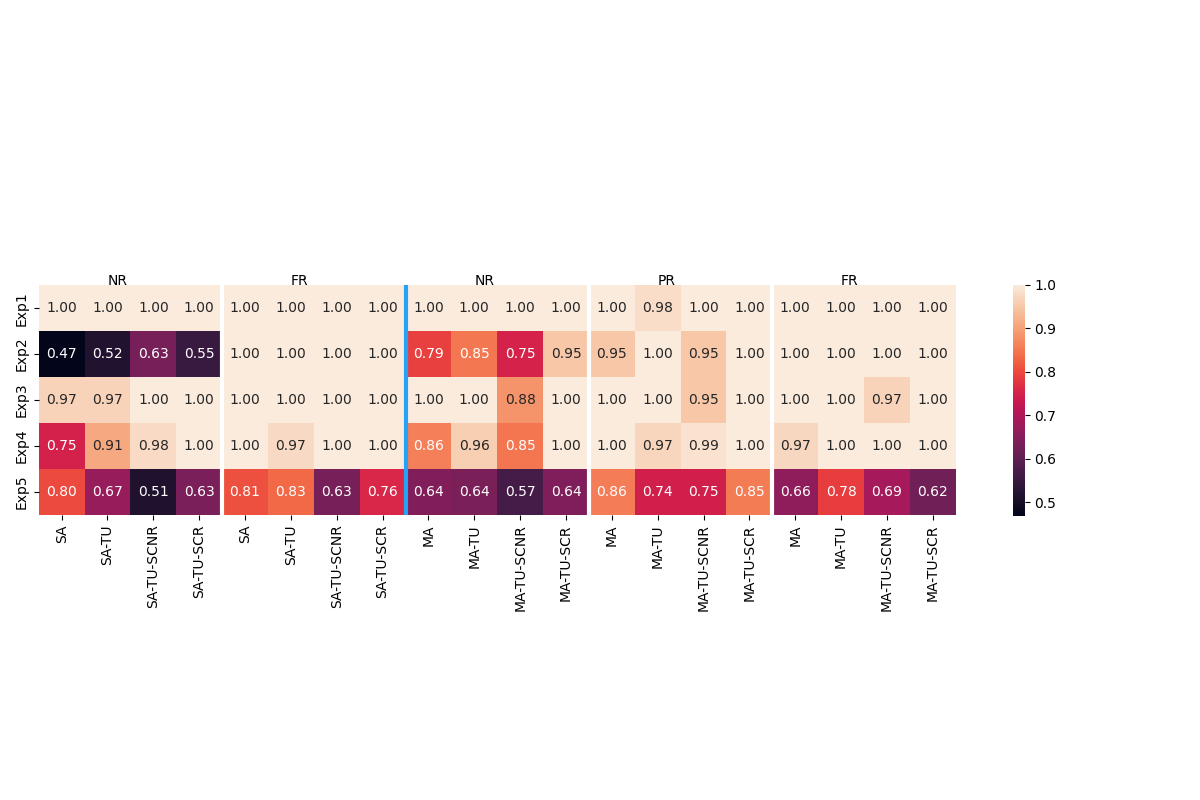}
\caption{Precision heatmap illustrating the performance of parameter step generation across different experiments. Higher values indicate better precision in predicting the correct parameters.}
\label{fig:false-color}
\end{figure}

\begin{figure}[htbp]
\centering
\includegraphics[width=1\linewidth, trim={0 4cm 0 5cm}, clip]{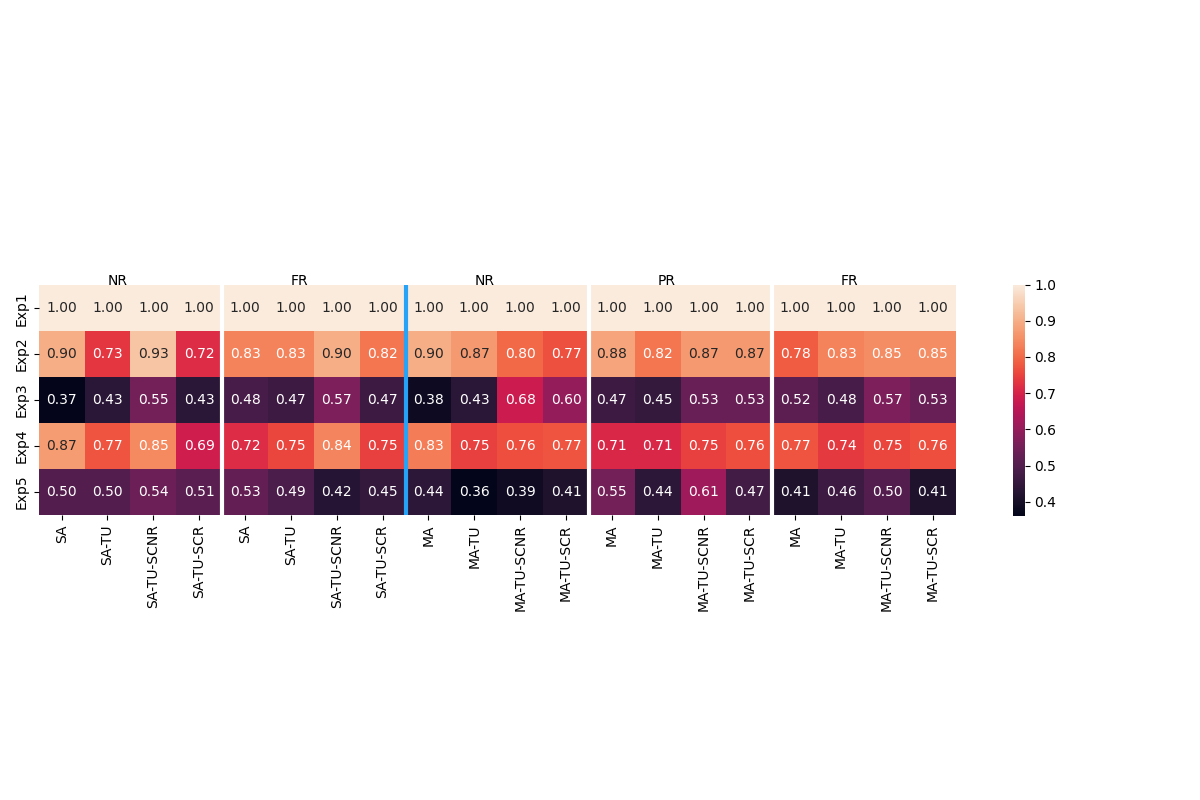}
\caption{Recall heatmap showing the performance of parameter step generation across different experiments. Higher values indicate better recall in identifying all the relevant parameters.}
\label{fig:false-color}
\end{figure}

\clearpage
\section{Experiment 5 False Positives by FR configurations}

This table compares the number of extra water addition steps generated by PR and FR multi-agent configurations. Higher values indicate more unnecessary water addition steps were generated compared to the ground truth.

\begin{table*}[!htb]
    \centering
    \begin{tabular}{lcc}
    \toprule
     \multicolumn{1}{l|}{} & \multicolumn{2}{c}{Number of Extra Water Steps} \\
     \multicolumn{1}{l|}{Experiment Type} &  PR & FR \\
    \midrule
    MA & 0 & 26 \\
    MA-TU & 0 & 24 \\
    MA-TU-GSC & 0 & 6 \\
    MA-TU-UGSC & 26 & 24 \\
    \bottomrule
    \end{tabular}
    \caption{Number of extra water addition steps generated by PR and FR configurations for Experiment 5.}
    \label{tab:placeholder}
\end{table*}

\clearpage
\section{Effect of RAG}
As discussed in the section 2.4.1 Reasoning Capacity is a Non-Negotiable Prerequisite for Quantitative Accuracy of the main text, the integration of the Retrieval-Augmented Generation (RAG) pipeline yielded notable improvements in F1 scores for both parameter generation and chemical generation steps across different experiments and configurations.
The following tables provide the corresponding breakdown of the F1 scores for parameter generation and chemical generation.

\begin{table}[!htb]

\begin{tabular}{lrrrrrrrrr}
\toprule
Configuration & experiment & \multicolumn{2}{r}{f1-no-rag} & \multicolumn{2}{r}{f1-rag} & \multicolumn{2}{r}{nrmse-no-rag} & \multicolumn{2}{r}{nrmse-rag} \\
 &  & mean & sem & mean & sem & mean & sem & mean & sem \\
\midrule
MA & 1 & 1.00 & 0.00 & 1.00 & 0.00 & 0.00 & 0.00 & 0.00 & 0.00 \\
MA & 2 & 0.71 & 0.03 & 0.75 & 0.04 & 0.04 & 0.04 & 0.05 & 0.04 \\
MA & 3 & 0.67 & 0.05 & 0.76 & 0.03 & 0.02 & 0.01 & 0.01 & 0.00 \\
MA & 4 & 0.79 & 0.04 & 0.82 & 0.05 & 0.07 & 0.02 & 0.12 & 0.03 \\
MA & 5 & 0.67 & 0.03 & 0.73 & 0.04 & 0.09 & 0.04 & 0.04 & 0.04 \\
\bottomrule
\end{tabular}
    \caption{Overall F1 scores.}
    \label{tab:placeholder}
\end{table}

\begin{table}[!htb]

\begin{tabular}{lrrrrrrrrr}
\toprule
Configuration & experiment & \multicolumn{2}{r}{F1-no-rag} & \multicolumn{2}{r}{f1-rag} & \multicolumn{2}{r}{nrmse-no-rag} & \multicolumn{2}{r}{nrmse-rag} \\
 &  & mean & sem & mean & sem & mean & sem & mean & sem \\
\midrule
MA & 1 & 1.00 & 0.00 & 1.00 & 0.00 & 0.00 & 0.00 & 0.00 & 0.00 \\
MA & 2 & 0.61 & 0.05 & 0.65 & 0.06 & 0.04 & 0.04 & 0.05 & 0.04 \\
MA & 3 & 0.68 & 0.06 & 0.78 & 0.04 & 0.02 & 0.01 & 0.01 & 0.00 \\
MA & 4 & 0.74 & 0.07 & 0.80 & 0.08 & 0.07 & 0.02 & 0.12 & 0.03 \\
MA & 5 & 0.82 & 0.01 & 0.82 & 0.04 & 0.09 & 0.05 & 0.04 & 0.04 \\
\bottomrule
\end{tabular}

    \caption{Chemical step generation F1 scores.}
    \label{tab:placeholder}
\end{table}

\begin{table}[!htb]
\begin{tabular}{lrrrrrrrrr}
\toprule
Configuration & experiment & \multicolumn{2}{r}{f1-no-rag} & \multicolumn{2}{r}{f1-rag} & \multicolumn{2}{r}{nrmse-no-rag} & \multicolumn{2}{r}{nrmse-rag} \\
 &  & mean & sem & mean & sem & mean & sem & mean & sem \\
\midrule
MA & 1 & 1.00 & 0.00 & 1.00 & 0.00 & 0.00 & 0.00 & 0.00 & 0.00 \\
MA & 2 & 0.87 & 0.03 & 0.91 & 0.00 & 0.04 & 0.04 & 0.05 & 0.04 \\
MA & 3 & 0.68 & 0.01 & 0.69 & 0.03 & 0.02 & 0.01 & 0.01 & 0.00 \\
MA & 4 & 0.86 & 0.01 & 0.85 & 0.01 & 0.07 & 0.02 & 0.12 & 0.03 \\
MA & 5 & 0.49 & 0.06 & 0.63 & 0.05 & 0.09 & 0.05 & 0.04 & 0.04 \\
\bottomrule
\end{tabular}
    \caption{Patameter step generation F1 scores.}
    \label{tab:placeholder}
\end{table}

\begin{table}[!htb]
\begin{tabular}{lrrrrrrrrr}
\toprule
exp\_type & experiment & \multicolumn{2}{r}{f1-no-rag} & \multicolumn{2}{r}{f1-rag} & \multicolumn{2}{r}{nrmse-no-rag} & \multicolumn{2}{r}{nrmse-rag} \\
 &  & mean & sem & mean & sem & mean & sem & mean & sem \\
\midrule
multi\_agent\_tools\_self\_checks & 1 & 1.00 & 0.00 & 0.99 & 0.01 & 0.00 & 0.00 & 0.00 & 0.00 \\
multi\_agent\_tools\_self\_checks & 2 & 0.85 & 0.03 & 0.84 & 0.02 & 0.00 & 0.00 & 0.08 & 0.05 \\
multi\_agent\_tools\_self\_checks & 3 & 0.82 & 0.03 & 0.75 & 0.05 & 0.01 & 0.00 & 0.05 & 0.02 \\
multi\_agent\_tools\_self\_checks & 4 & 0.90 & 0.01 & 0.86 & 0.02 & 0.01 & 0.00 & 0.03 & 0.01 \\
multi\_agent\_tools\_self\_checks & 5 & 0.75 & 0.04 & 0.76 & 0.04 & 0.06 & 0.06 & 0.06 & 0.06 \\
\bottomrule
\end{tabular}
    \caption{Overall F1 scores.}
    \label{tab:placeholder}
\end{table}

\begin{table}[!htb]
\begin{tabular}{lrrrrrrrrr}
\toprule
exp\_type & experiment & \multicolumn{2}{r}{f1-no-rag} & \multicolumn{2}{r}{f1-rag} & \multicolumn{2}{r}{nrmse-no-rag} & \multicolumn{2}{r}{nrmse-rag} \\
 &  & mean & sem & mean & sem & mean & sem & mean & sem \\
\midrule
multi\_agent\_tools\_self\_checks & 1 & 1.00 & 0.00 & 1.00 & 0.00 & 0.00 & 0.00 & 0.00 & 0.00 \\
multi\_agent\_tools\_self\_checks & 2 & 0.80 & 0.05 & 0.79 & 0.03 & 0.00 & 0.00 & 0.08 & 0.05 \\
multi\_agent\_tools\_self\_checks & 3 & 0.87 & 0.05 & 0.78 & 0.08 & 0.01 & 0.00 & 0.05 & 0.02 \\
multi\_agent\_tools\_self\_checks & 4 & 0.93 & 0.02 & 0.89 & 0.02 & 0.01 & 0.00 & 0.03 & 0.01 \\
multi\_agent\_tools\_self\_checks & 5 & 0.91 & 0.03 & 0.88 & 0.06 & 0.06 & 0.06 & 0.06 & 0.06 \\
\bottomrule
\end{tabular}
    \caption{Chemical step generation F1 scores.}
    \label{tab:placeholder}
\end{table}

\begin{table}[!htb]
\begin{tabular}{lrrrrrrrrr}
\toprule
exp\_type & experiment & \multicolumn{2}{r}{f1-no-rag} & \multicolumn{2}{r}{f1-rag} & \multicolumn{2}{r}{nrmse-no-rag} & \multicolumn{2}{r}{nrmse-rag} \\
 &  & mean & sem & mean & sem & mean & sem & mean & sem \\
\midrule
multi\_agent\_tools\_self\_checks & 1 & 1.00 & 0.00 & 0.99 & 0.01 & 0.00 & 0.00 & 0.00 & 0.00 \\
multi\_agent\_tools\_self\_checks & 2 & 0.92 & 0.02 & 0.90 & 0.02 & 0.00 & 0.00 & 0.08 & 0.05 \\
multi\_agent\_tools\_self\_checks & 3 & 0.69 & 0.02 & 0.67 & 0.00 & 0.01 & 0.00 & 0.05 & 0.02 \\
multi\_agent\_tools\_self\_checks & 4 & 0.86 & 0.01 & 0.84 & 0.02 & 0.01 & 0.00 & 0.03 & 0.01 \\
multi\_agent\_tools\_self\_checks & 5 & 0.57 & 0.06 & 0.65 & 0.04 & 0.06 & 0.06 & 0.06 & 0.06 \\
\bottomrule
\end{tabular}
    \caption{Parameter step generation F1 scores.}
    \label{tab:placeholder}
\end{table}

\clearpage
\section{Spearman correlation}

The following list presents the alternative, scientifically valid step sequence for experiment 3. This sequence was used as the ground-truth for the Spearman correlation analysis shown in the bottom panel of Figure 11 in the main text. 



\begin{enumerate}[itemsep=0pt]
    \item Add benzaldehyde (ul) to vials in Plate 1
    \item Add 1-bromobutane (ul) to vials in Plate 1
    \item Add 1-iodobutane (ul) to vials in Plate 1
    \item Add 1-chlorobutane (ul) to vials in Plate 1
    \item Add 3-bromopropene (ul) to vials in Plate 1
    \item Add benzyl bromide (ul) to vials in Plate 1
    \item Add 3-bromobut-1-ene (ul) to vials in Plate 1
    \item Add 3-bromobut-2-ene (ul) to vials in Plate 1
    \item Add 2-bromoethyl cyanide (ul) to vials in Plate 1
    \item Add aqueous ammonia (ul) to vials in Plate 1
    \item Add water (ul) to vials in Plate 1
    \item Set Cap for vials in Plate 1
    \item Set StirRate to 700 rpm in Plate 1
    \item Set HeatingTemp to 60 C in Plate 1
    \item Set Delay to 480 min in Plate 1
    \item Set HeatingTemp to 25 C in Plate 1
    \item Set StirRate to 0 rpm in Plate 1
\end{enumerate}